\documentclass[letterpaper]{article} % DO NOT CHANGE THIS
\usepackage{aaai24}  % DO NOT CHANGE THIS
\usepackage{times}  % DO NOT CHANGE THIS
\usepackage{helvet}  % DO NOT CHANGE THIS
\usepackage{courier}  % DO NOT CHANGE THIS
\usepackage[hyphens]{url}  % DO NOT CHANGE THIS
\usepackage{graphicx} % DO NOT CHANGE THIS
\usepackage{natbib}  % DO NOT CHANGE THIS AND DO NOT ADD ANY OPTIONS TO IT
\usepackage{caption} % DO NOT CHANGE THIS AND DO NOT ADD ANY OPTIONS TO IT
\usepackage[linesnumbered,ruled]{algorithm2e}
\usepackage[noend]{algpseudocode}
\usepackage{eqparbox}

\newcommand{\WRP}{\par\qquad\(\hookrightarrow\)\enspace}
\SetAlgorithmName{Algorithm}{Algorithm}{Algorithm}
\IncMargin{0.5em}
\SetCommentSty{textnormal}
\SetNlSty{}{}{:}
\SetAlgoNlRelativeSize{0}
\SetKwInput{KwGlobal}{Global}
\SetKwInput{KwPrecondition}{Precondition}
\SetKwProg{Proc}{Procedure}{:}{}
\SetKwProg{Func}{Function}{:}{}
\SetKw{And}{and}
\SetKw{Or}{or}
\SetKw{To}{to}
\SetKw{DownTo}{downto}
\SetKw{Break}{break}
\SetKw{Continue}{continue}
\SetKw{SuchThat}{\textit{s.t.}}
\SetKw{WithRespectTo}{\textit{wrt}}
\SetKw{Iff}{\textit{iff.}}
\SetKw{MaxOf}{\textit{max of}}
\SetKw{MinOf}{\textit{min of}}
\SetKwBlock{Match}{match}{}{}

\usepackage{enumitem}
\usepackage{multirow}
\usepackage{adjustbox}
\usepackage{floatrow}
\usepackage{amsmath, nccmath}
\usepackage{amssymb}
\usepackage{mathtools}
\usepackage{amsthm}
\newfloatcommand{capbtabbox}{table}[][\FBwidth]
\DeclareMathOperator{\E}{\mathbb{E}}
\DeclareMathOperator*{\argmax}{arg\,max}
\theoremstyle{definition}
\newtheorem{definition}{Definition}
\usepackage[capitalize,noabbrev]{cleveref}

\usepackage{newfloat}
\usepackage{listings}
\DeclareCaptionStyle{ruled}{labelfont=normalfont,labelsep=colon,strut=off} % DO NOT CHANGE THIS
\floatstyle{ruled}
\newfloat{listing}{tb}{lst}{}
\floatname{listing}{Listing}
\title{MaxEnt Loss: Constrained Maximum Entropy \\ for Calibration under Out-of-Distribution Shift}
\author {
    Dexter Neo,
    Stefan Winkler, 
    Tsuhan Chen \\
}
\affiliations {
    School of Computing, National University of Singapore\\
    {e0534450@u.nus.edu, \{winkler, tsuhan\}@nus.edu.sg}
}

\usepackage{bibentry}
\begin{document}
\maketitle
\begin{abstract}
We present a new loss function that addresses the out-of-distribution (OOD) network calibration problem. While many objective functions have been proposed to effectively calibrate models in-distribution (ID), our findings show that they do not always fare well OOD. Based on the Principle of Maximum Entropy, we incorporate helpful statistical constraints observed during training, delivering better model calibration without sacrificing accuracy. We provide theoretical analysis and show empirically that our method works well in practice, achieving state-of-the-art calibration on both synthetic and real-world benchmarks. Our code is available at $\underline{\textit{https://github.com/dexterdley/MaxEnt-Loss}}$.
\end{abstract}

\section{Introduction}
Recent advances in machine learning have given rise to large neural networks with strong recognition performance in fields such as computer vision and natural language processing. Neural networks are increasingly used in areas where safety is a concern, such as self-driving cars \cite{Bojarski2016EndTE}, medical prognosis \cite{Esteva2017DermatologistlevelCO, bandi2018detection} and facial expression recognition \cite{vonikakis2021morphsetaugmenting, Neo_2023_CVPR, neo2023ferc}. Despite their popularity, deep neural networks have a tendency to be poorly calibrated. 

\begin{figure}[!tb]
\centering
\includegraphics[width=\columnwidth]{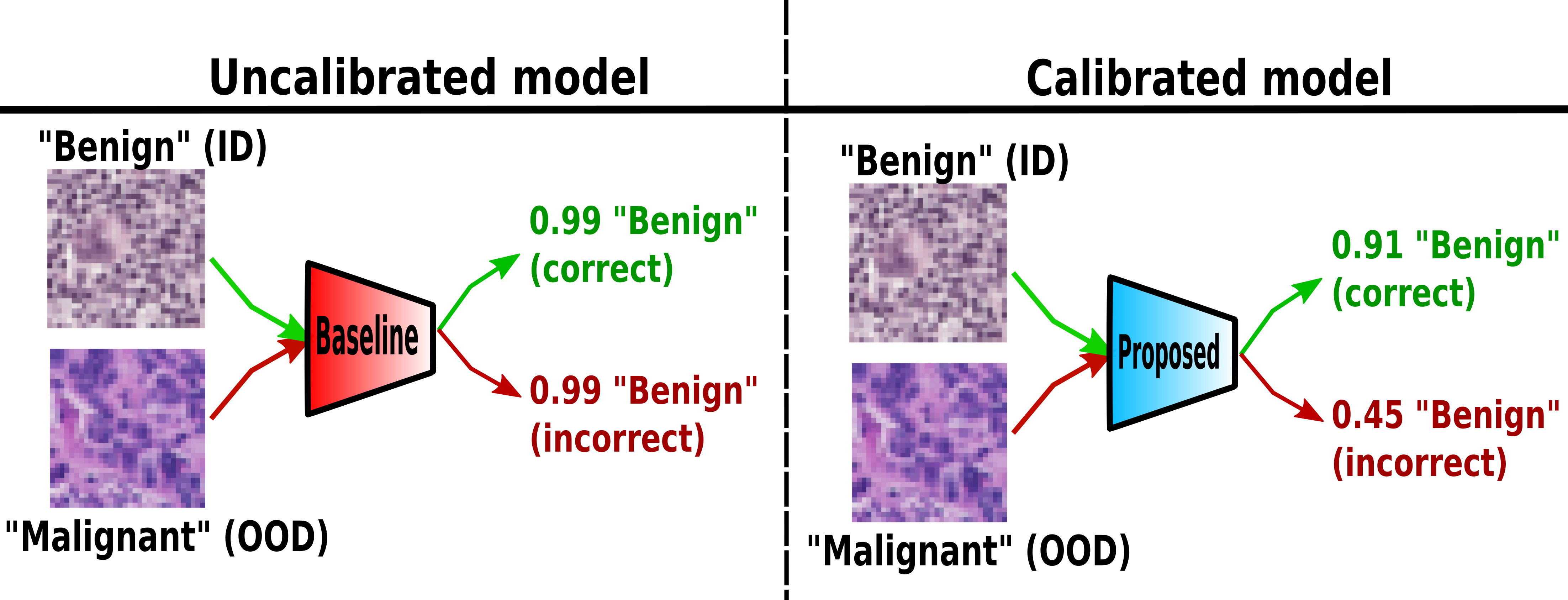}
\caption{Uncalibrated models (left) make overconfident OOD misdiagnoses, resulting in dire consequences. Well-calibrated models (right) exhibit lower confidence, reflecting their uncertainty for OOD samples.}
\label{fig:fig1}
\end{figure}

Calibration refers to the model's correctness with regard to its predicted probabilities. In other words, models tend to overconfidently misclassify samples and erroneously recognize correct classes with low confidence. This often creates mistrust due to the mismatch between model correctness and confidence. For severe cases, uncalibrated models can cause harm, resulting in serious consequences such as overconfidently misidentifying cancerous cells (see \cref{fig:fig1}). This is especially common when uncalibrated models encounter samples that are out-of-distribution (OOD) from the training set. Ideally, a well-calibrated classifier should behave unconfidently and predict low probabilities whenever it misclassifies samples. Furthermore, we would like our models to not only remain accurate and well-calibrated in-distribution (ID) but to also provide further robustness against OOD shifts for safe deployment \cite{Thulasidasan2019OnMT, verified_uncertainty}.

The current hypothesis as to why modern neural networks are miscalibrated is that these large models with millions of parameters have the capacity to learn and overfit to the given training data \cite{pmlr-v70-guo17a}. This is especially true if model training is done using the cross-entropy (CE) loss, since the CE loss can only be fully minimized when probabilities $\hat{P}$ are equal to the one-hot ground truth $y$. This means that even though the accuracy is at 100\%, CE loss can still be positive and minimized further by increasing the confidence of the predicted probabilities, resulting in overfitting and miscalibration. While many techniques have been proposed to address calibration, a category of methods include objective functions that pair CE loss with auxiliary terms. Specifically, objective functions such as Focal, Inverse Focal and Poly loss \cite{focal, wang2021rethinking, leng2022polyloss} add an additional penalty term to control the confidence of predictions. Other objective functions such as AvUC and Soft-AvUC loss introduce a differentiable utility term based on accuracy and uncertainty \cite{krishnan2020improving, karandikar2021soft}.

Contrary to these methods, we follow the work of \citet{mukhoti2020calibrating} and further explore the Principle of Maximum Entropy (MaxEnt) \cite{MaxEnt}. We propose a novel regularization technique in the form of a general loss function for improving calibration based on constrained maximum entropy. Our method works by introducing additional constraints that complement loss functions typically used in supervised learning. We provide systematic comparisons of model accuracy and calibration for image classification tasks. Our main contributions can be summarized as follows:
\begin{enumerate}%[noitemsep,nolistsep]
    %\small %cannot change
    \item \textbf{MaxEnt Loss:} We explore the theoretical relations between the Principle of Maximum Entropy and Focal loss, based on which we propose a novel loss function with three different forms, showing how constraints can be introduced to further improve calibration.
    \item \textbf{Automated Hyperparameter Tuning:} Based on the constraints, our framework provides an automated estimate of the optimal Lagrange multipliers, with no need for manual tuning.
    \item \textbf{Evaluation on OOD shifts:} Our experiments show that MaxEnt loss remains robust in terms of accuracy and calibration for both synthetic and in-the-wild distribution shift benchmarks. We also analyze the ordering of the model's feature norms under increasingly shifted inputs.
    \item \textbf{Ad-hoc calibration:} Our method is non-restrictive and works well in combination with popular ad-hoc calibration methods such as temperature scaling and label smoothing.
\end{enumerate}

\section{Related Work}
\textbf{Network Calibration:} Existing methods for calibrating neural networks can be categorized roughly as follows: (1) Methods that approximate the true joint distribution or latent hidden vector $z$ using generative models such as VAE \cite{Kingma2014AutoEncodingVB}, Cycle-GAN \cite{CycleGAN2017}. (2) Methods that directly regularize the output probabilities such as isotonic regression \cite{isotonic_regression}, Bayesian binning \cite{10.5555/2888116.2888120}, splines \cite{gupta2021calibration}, objective functions \cite{krishnan2020improving, karandikar2021soft, wang2021rethinking, leng2022polyloss, Cheng_2022_CVPR} and temperature scaling methods \cite{pmlr-v70-guo17a, Kull2019BeyondTS}. A recent summary of model calibration can be found in \cite{minderer2021revisiting}.

\noindent \textbf{Calibration under OOD shift:} For OOD problems, test inputs do not align with the training set \cite{Ovadia2019CanYT}. This phenomenon can be caused by either (1) completely OOD test inputs that belong to an OOD class not from the ground truth labels \cite{Du2022VOSLW}, or (2) shifted OOD test inputs caused by perturbations and corruptions \cite{Hendrycks2017ABF}. Apart from the loss functions introduced previously, many other strategies have been proposed to tackle OOD calibration. This includes methods such as multi-domain temperature scaling \cite{yu2022robust}, sampling Gaussians for domain drifts \cite{Tomani_2021_CVPR} and transferable calibration \cite{ximeiwang}. For this work, we mainly focus on loss functions for on-the-fly OOD calibration for both synthetic and in-the-wild data, rather than pre- or post-hoc techniques. For a recent review of OOD shifts, we refer to \cite{Wiles2022AFA}. % for a summary of the literature. 

\noindent \textbf{Maximum Entropy:} \citet{pereyra2017regularizing} have shown that directly penalizing neural networks with the maximum entropy term helps prevent overconfidence, resulting in better generalization. The Principle of Maximum Entropy \cite{MaxEnt} has a long standing in information theory where we maximize the model's entropy subject to constraints derived from the training set \cite{berger-etal-1996-maximum}. Focal loss \cite{focal} was originally proposed for object detection, yet it can also be used for improving calibration. Mathematically, Focal loss is a general form of CE loss with an additional entropy term \cite{mukhoti2020calibrating} -- reducing Focal loss simultaneously \textit{minimizes} the KL divergence and \textit{maximizes} the entropy, discouraging overconfidence. The MaxEnt method is also used in other tasks, such as Fine-Grained Visual Classification  \cite{Dubey2018MaximumEntropyFC}, where classes may be visually similar, and reinforcement learning \cite{Haarnoja2018SoftAO, Neo2023DSACCCM}, where high entropy policies tend to encourage stochasticity and improve exploration . 

\section{Preliminaries}
Consider a classification task over a dataset $D$ with $N$ number of samples $(x_i, y_i)^N_{i=1}$, where X, Y denote input feature and label space, and $\mathcal{Y} = [1, 2, ..., K]$ is a fixed array containing all $K$ class indices. Given an arbitrary input datum $x_i$, the task is modelled by a neural network with learnable parameters $\boldsymbol{\theta}$ and a penultimate layer containing $K$ neurons, which output logits $g^\theta_i(x)$. The model learns to estimate the posterior distribution which are a set of valid probabilities such that $\sum^K_{k=1} P_i(y_k|x) = 1$, after the softmax function $\frac{\exp{g^\theta_i(x)}}{ \sum^K_{k=1} \exp{g^\theta_k(x)}}$. The predicted top-1 class is then simply $\hat{y} := \argmax g^\theta_i(x)$, with the corresponding confidence score $\hat{P}:= \max P_i(y_k|x)$. In theory, a model is considered perfectly calibrated iff the model's probabilities match the true posterior distribution, satisfying the definition $\mathbb{P}(\hat{y}=y| \hat{P}=P ) = P \quad \forall \in P[0-1]$. Realistically, achieving this level of calibration is infeasible as the true posterior distribution remains unknown. In the following we list a few error metrics have been proposed to approximate calibration; for the definitions of additional calibration metrics and results please refer to the Appendix.

\noindent \textbf{Expected Calibration Error (ECE):}  Calibration error is commonly estimated using ECE \cite{10.5555/2888116.2888120}. It is computed by splitting the model's probabilities into $B$ bins. Let $n_b$, \textrm{acc} and \textrm{conf} represent the number of samples, average accuracy and confidence for each partitioned bin. The weighted absolute differences between \textrm{acc} and \textrm{conf} for each bin is calculated using the following formula: ECE $= \sum^B_{b=1} \frac{n_b}{N} | \textrm{acc}(b) - \textrm{conf}(b)|$.

\noindent \textbf{Classwise ECE (CECE):}  ECE only considers the $\max$ confidence of the predicted probabilities. In certain scenarios, one would also require the probabilities of all other classes to be well calibrated, therefore CECE is proposed as an simple extension of ECE, considering all K classes \cite{Nixon2019MeasuringCI}: CECE $= \frac{1}{K} \sum^B_{b=1} \sum^{K}_{k=1} \frac{n_{b,k}}{N} | \textrm{acc}(b,k) - \textrm{conf}(b,k) |$.

\noindent \textbf{Kolmogorov-Smirnov Error (KSE):} The approximation of calibration errors often requires the histograms/binning of empirical distributions. This causes an over-reliance on binning, which is sensitive to the number of bins chosen. Inspired by the Kolmogorov-Smirnov test, \cite{gupta2021calibration} propose a numerical approximation of the equality between two cumulative distributions without the need for binning. Using the authors' notation, the model's probabilities are abbreviated as $z_k$ with the integral form of the Kolmogorov-Smirnov error for top-1 classification defined as: KSE $= \int^1_0 |P(k|z_k) - z_k| P(z_k) dz_k $.  % #\cref{extra_metrics}.

\begin{figure}[tb]
\centering
\includegraphics[width=\columnwidth]{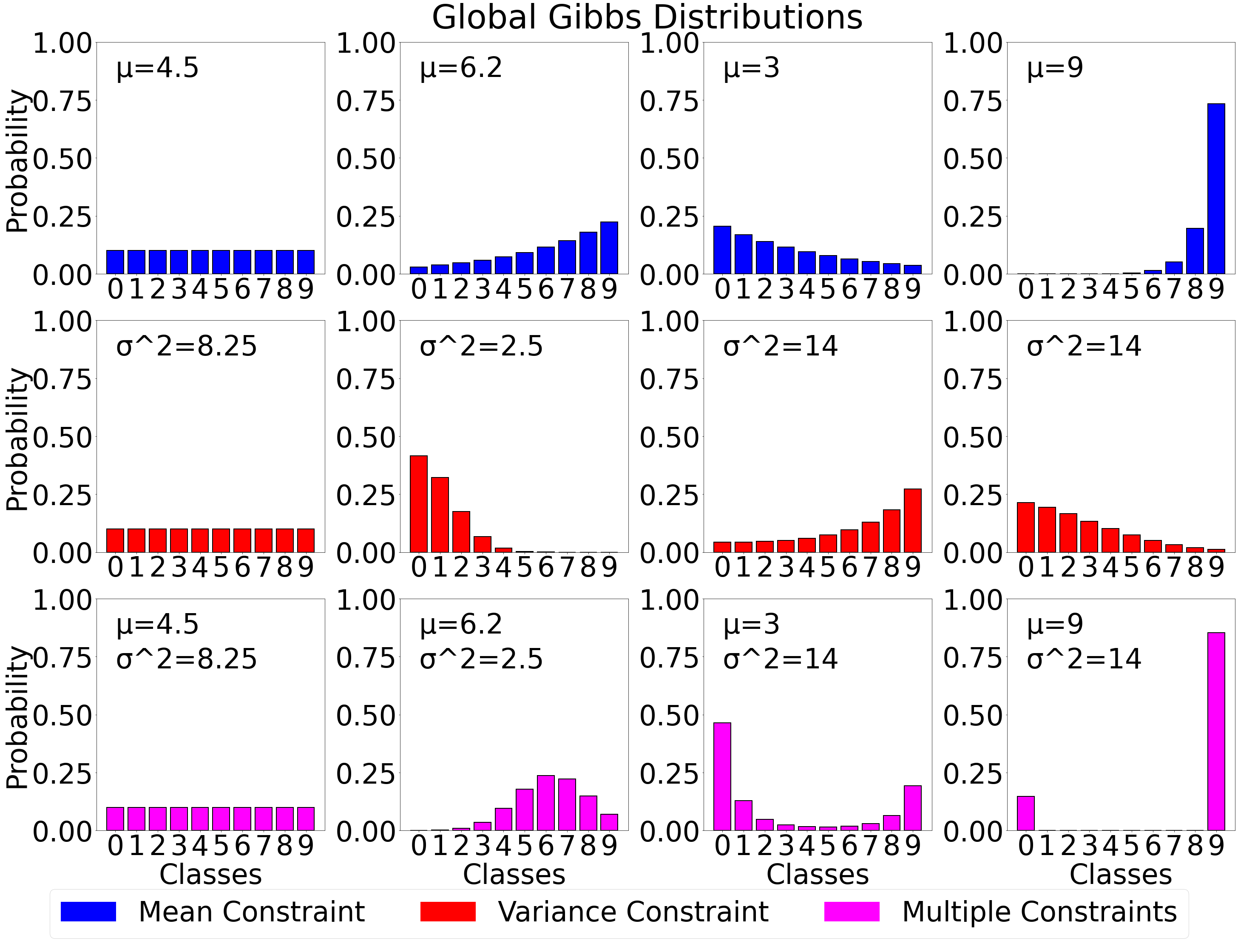}
\caption{In the face of unknown OOD, we argue that predictions should not deviate too far from the observed global Gibbs distribution. For different values of constraints, higher $\mu$ values tend toward larger classes, while a higher $\sigma^2$ results in distributions being more spread out.}
\label{fig:simple_maxent}
\end{figure}

\section{Methodology} %Methods
%In the face of OOD, we propose that predictions should not deviate too far from the global Gibbs distribution observed in the training set. Different expectations result in different values of constraints, higher $\mu$ values tend toward larger classes, while a higher $\sigma^2$ results in distributions being more spread out.
In this section, we show the relationship between  Maximum Entropy \cite{MaxEnt}, confidence penalty term \cite{pereyra2017regularizing} and Focal loss \cite{focal}. We also show the effects of different constraints placed on the MaxEnt method and describe how they can be developed into a novel objective function for model calibration that we call MaxEnt Loss.

\subsection{Principle of Maximum Entropy}
The Principle of Maximum Entropy\footnote{~We refer readers to \cite{MaxEnt} along with examples included in the Appendix for an extended background of the Principle of Maximum Entropy.} is a probability distribution that maximizes the Shannon entropy subject to the given inequality constraints. Specifically, the general discrete form of the MaxEnt method is given by:
\begin{equation}
\begin{split}
&\max H(P_i(y_k|x)) =  - \sum^{K}_{k=1} P_i(y_k|x) \log P_i(y_k|x) \\
&\text{subject to} \sum_k P_i(y_k|x) f_n(\mathcal{Y}) \le c_n \text{ for all constraints} f_n(\mathcal{Y})
\end{split}
\label{max_ent_method}
\end{equation}
where $f_n(\mathcal{Y})$ is a function of the random variable vector $\mathcal{Y}$ and the respective Lagrange multipliers $\lambda_n$ for each $n$ number of constraints $c_n$. In theory, the MaxEnt method allows for any arbitrary function $f_n(\mathcal{Y})$ to be computed within the expectation, in practice we can simply assign a fixed vector $\mathcal{Y}$ containing the class labels as our function. 

Following \cref{max_ent_method}, we build a simple MaxEnt model using a ten-class example and illustrate the effects of different values of $c_n$ in \cref{fig:simple_maxent}. We show the global Gibbs distribution as a function of the given random variable and subjected constraints. The constraints are scalars computed from the expectation of the random variable such as $\E[\mathcal{Y}]$ and $\E[\mathcal{Y}^2]$. For example, if the class label frequency/prior distribution $P(\mathcal{Y})$ of the training set is uniform, then the expected average $\mu$ would be $\sum^{K=10}_{k=1} \mathcal{Y}P(\mathcal{Y})=4.5$, with the corresponding expected variance $\sigma^2$ given as $\sum^{K=10}_{k=1} \mathcal{(Y - \mu)}^2 P(\mathcal{Y})=8.25$ (see column 1 of \cref{fig:simple_maxent}). In row 1, increasing $\mu>4.5$ results in a shift towards larger classes and increasing $\sigma^2>8.25$ will give a distribution with a higher variance. For row 3, when both constraints are combined, the distributions shift to jointly accommodate both the mean and variance.

\subsection{Relation to Focal Loss}
Next, we show the relation between the Principle of Maximum Entropy and Focal loss, which reduces the emphasis on easily classified samples \cite{focal}. Consider the multi-class form of Focal loss where $\mathcal{L}_F = -\sum_k \bigl(1 - P_i(y_k|x_i) \bigr)^\gamma \log P_i(y_k|x_i)$, with the hyperparameter $\gamma \ge 0$. By setting $\gamma=1$, $\mathcal{L}_F$ can be expanded and re-written as the CE loss with a confidence penalty term (Shannon's entropy):
\begin{equation}
\mathcal{L}_{F} = -\sum_k \underbrace{\log P_i(y_k|x)}_{\text{CE Loss}} - \underbrace{H(P_i(y_k|x) ) }_{\text{Shannon term}}
\label{focal_loss}
\end{equation}
Supposedly, even if other values of $\gamma>1$ are chosen, \cref{focal_loss} still holds true such that the Shannon term is paired with a polynomial. This additional entropy term is useful for preventing peaked distributions and delivers better generalization \cite{mukhoti2020calibrating}. 

Connecting \cref{max_ent_method} and \cref{focal_loss}, we argue that maximizing the model's entropy subject to constraints computed from the prior knowledge observed would be a possible approach for OOD scenarios. Since it is impossible to know beforehand the type or intensity of OOD shift, utilizing any additional information during training can be useful OOD. For example, if the class label frequencies of the training set are uniform, we 
should expect the expectations of the classifier's predictions to be closer to that of a uniform distribution, especially if the inputs are progressively shifted from the training samples.

\subsection{MaxEnt Loss for End-to-End Training} %(Proposed)
Given the above, we now demonstrate how constraints can be added to the Focal loss. We propose three forms of MaxEnt Loss for end-to-end model calibration:

\begin{algorithm*}[!htb]
\SetAlgoNoLine
\SetAlgoNoEnd
\DontPrintSemicolon
\KwData{Given training set $D = (x_i, y_i)^N_{i=1}$ }
\SetKwFunction{fOptimize}{Optimize}
\SetKwFunction{fLabelFusion}{LabelFusion}
\SetKwFunction{fNewtonRaphson}{NewtonRaphson}
\BlankLine
Initialize neural network parameters $\boldsymbol{\theta}$ and learning rate schedule $\alpha$ \;
Compute the global and local expectations for the mean and variance constraints $\mu, \sigma^2$ \WRP $\E[\mathcal{Y}] = \mu$ and $\E[\mathcal{Y}^2] = \sigma^2 $\;
Solve numerically for $\lambda_n \leftarrow \fNewtonRaphson{}$ \tcp*{Use a root-finder to obtain $\lambda_n$}
\BlankLine
\lFor{$e \in epochs$}{\;
\hskip2.0em \lFor{$i \in B$}{ \tcp*{Sample mini-batch of size $B$}
\BlankLine
\hskip4.0em Calculate MaxEnt Loss: $\mathcal{L}_{ME} = \mathcal{L}_{F} + \sum^M_{n=1}  \lambda_n \left( \sum^K_{k=1} \mathcal{Y} P_i(y_k|x_i) - c_n \right) $ \;
\hskip4.0em $\boldsymbol{\theta} \leftarrow \boldsymbol{\theta} - \alpha \Delta \mathcal{L}_{ME} $ \tcp*{Update parameters $\boldsymbol{\theta}$ by gradient descent}
\Return $\boldsymbol{\theta}$
}
}
%Mixed label extraction function
\Func{\fNewtonRaphson{}}{
$\delta$ = 1e-15 \tcp*{A small tolerance or stopping condition}
\While{ g($\lambda$) $> \delta$}{
$\lambda_{n+1} = \lambda_n -\frac{g(\lambda)}{g'(\lambda)} $ \tcp*{Update Lagrange Multipliers $\lambda_n$}
}
\Return $\lambda_n$
}
\caption{Constrained MaxEnt Loss Optimization }
\label{alg:algorithm}
\end{algorithm*}

\begin{definition}
[Mean Constraint] Consider the expected average $\mu$ of the target distribution to be constrained, we can add the following mean constraint terms to \cref{focal_loss} and aim to minimize the following objective function:
\end{definition}
\begin{equation}
\begin{split}
&\mathcal{L}_{ME}^M = -\sum_k \log P_i(y_k|x) - H(P_i(y_k|x) )  \\ 
&+ \lambda_\mu \Biggl[ \underbrace{\sum_k \mathcal{Y} P_i(y_k|x) - \mu_G}_{\text{Global mean constraint}}  + \underbrace{\sum_k \mathcal{Y} P_i(y_k|x) - \mu_{Lk} }_{\text{Local mean constraint}} \Biggr]
\label{expo_final}
\end{split}
\end{equation}
where $\mu_G$ is the global expected average $\E[\mathcal{Y}]$ computed from the prior distribution described earlier, and $\mu_{Lk}$ is the local expectation for the $k$th class, which can be computed from the target labels $\sum_k \mathcal{Y}y_k = \mu_{Lk}$ and $\lambda_\mu$ is the Lagrange multiplier for the mean constraint form.

\begin{definition}
[Variance Constraint] Next, consider the case where the expected variance $\sigma^2$ of the target distribution is to be constrained, the variance constraints can be added to \cref{focal_loss}:
\end{definition}
\begin{equation}
\begin{split}
&\mathcal{L}_{ME}^V = -\sum_k \log P_i(y_k|x) - H(P_i(y_k|x) )  \\ 
&+ \lambda_{\sigma^2} \Biggl[ \underbrace{  \sum_k \mathcal{Y}^2 P_i(y_k|x) - \sigma^2_G }_{\text{Global variance constraint}} + \underbrace{\sum_k \mathcal{Y}^2 P_i(y_k|x) - \sigma^2_{Lk}}_{\text{Local variance constraint}} \Biggr]
\label{gaussian_final}
\end{split}
\end{equation}
where $\sigma^2_G$ is the global expected variance $\E[\mathcal{Y}^2]$ and $\sum_k \mathcal{Y}^2 y_k = \sigma^2_{Lk}$ is the expected local variance for the $k$th class with $\lambda_{\sigma^2}$ as the corresponding Lagrange multiplier. For this form, we assume that there is no knowledge of the expected average/mean constraint.

\begin{definition}
[Mean and Variance Constraints] Finally, when both the expected average and variance of the target distribution are to be considered, we can combine both constraints. The objective function for this form is given by:
\end{definition}
\begin{equation}
\small
\begin{split}
& \mathcal{L}_{ME}^{M+V} = \mathcal{L}_F +\lambda_\mu \Biggl[ \sum_k \mathcal{Y} P_i - \mu_G + \sum_k \mathcal{Y} P_i - \mu_{Lk} \Biggr] \\
& + \lambda_{\sigma^2} \Biggl[ \sum_k (\mathcal{Y} - \mu)^2 P_i - \sigma^2_G + \sum_k (\mathcal{Y} - \mu)^2 P_i - \sigma^2_{Lk} \Biggr]
\label{multiple_final}
\end{split}
\end{equation}

For this case, the variances are computed together with the expected average $\sum_k (\mathcal{Y} - \mu)^2 y_k = \sigma^2_{Lk}$, and the Lagrange multipliers $\lambda_\mu$ and $\lambda_{\sigma^2}$ need to be solved simultaneously. For all three definitions, the respective Lagrange multipliers $\lambda_n$ can be solved cheaply using traditional numerical root-finders that require only CPU. In our work, we select Newton Raphson's method as our root-finder which utilizes a helper function $g(\lambda)$ and its derivative $g'(\lambda)$ to solve for $\lambda_n$ in $\mathcal{O}(n)$ time.
This step is described in lines 4 and 11-15 of \cref{alg:algorithm}, along with more details on the rest of our method. 

We highlight that global expected mean and variance are computed from the prior distribution $P(y_k)$; the local mean and variances are computed for each class label. For example, in the case of one-hot labels the local averages are always $\mu_{Lk} = y_k$, however this is no longer true if the ground truth labels are not one-hot (e.g.\ label smoothing). 

After the constraints are computed, each of the Lagrange multipliers $\lambda_n$ are solved numerically before training is performed. The Appendix contains examples and proofs on this computation for single and multiple constraints. In our ablation study, we also include a discussion regarding the empirical effects of local constraints.

\begin{figure*}[!htb]
\small
\centering
\includegraphics[width=\textwidth]{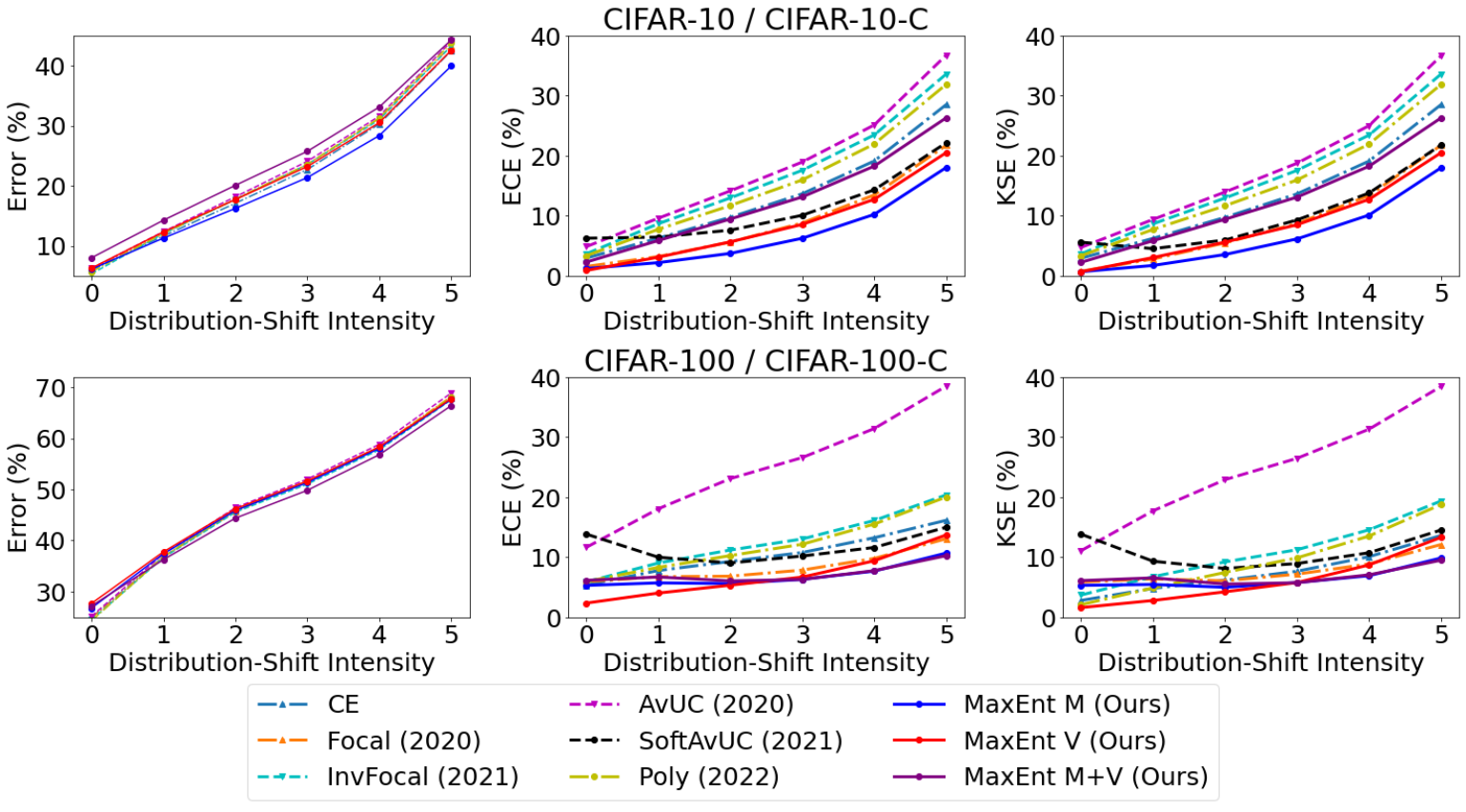}
%\vspace{-6mm}
\caption{Test and calibration error curves highlighting the performance of different loss functions on CIFAR/CIFAR-C. As distribution shifts worsen from 0 to 5, all methods converge to similar test errors, while our method remains well calibrated.}
\label{fig:cifar_plots}
\end{figure*}

\section{Experiments and Results}
We perform experiments on six popular OOD shift image classification benchmarks and evaluate our method against recently proposed calibration objective functions. Specifically, we compare against the following baseline losses: CE, Focal, Inverse Focal \cite{wang2021rethinking}, AvUC \cite{krishnan2020improving}, Soft AvUC \cite{karandikar2021soft} and Poly loss \cite{leng2022polyloss}. 

For our analysis on synthetic OOD, we use ResNet-18 and  ResNet-50 \cite{He2016DeepRL} with SGD optimizer for CIFAR and TinyImageNet, respectively. For in-the-wild OOD, we use  ResNet-18 and DenseNet-121 \cite{Huang2017DenselyCC} with Adam optimizer \cite{Kingma2015AdamAM}. Additional details about the OOD corruptions, experimental setup, and hyperparameters are provided in the Appendix. We show examples of each dataset in \cref{fig:ood_figures} and describe the details of the following tasks.

\noindent \textbf{Synthetic OOD}: We make use of standard benchmarks for CIFAR10/CIFAR100/TinyImageNet and their corrupted forms CIFAR10-C/CIFAR100-C/TinyImageNet-C.
\begin{enumerate}%[noitemsep,nolistsep]
    \item CIFAR10/CIFAR100 \cite{Krizhevsky09} contains RGB colored images (32x32) with ten or hundred classes. 45,000\slash 5,000\slash 10,000 images for training/validation/testing.
    \item TinyImagenet \cite{imagenet} is a subset of ImageNet with 200 classes, with images of size 64x64. 100,000 for training and 10,000 for validation/testing.
    \item CIFAR10-C/CIFAR100-C/TinyImagenet-C \cite{hendrycks2019robustness} The corrupted form of CIFAR and TinyImagenet, comprising a total of 19 different transformations, with the initial 10,000 images of severity level one and the last 10,000 images of severity five.
\end{enumerate}

\begin{figure}[!htb]
\centering
\includegraphics[width=\columnwidth]{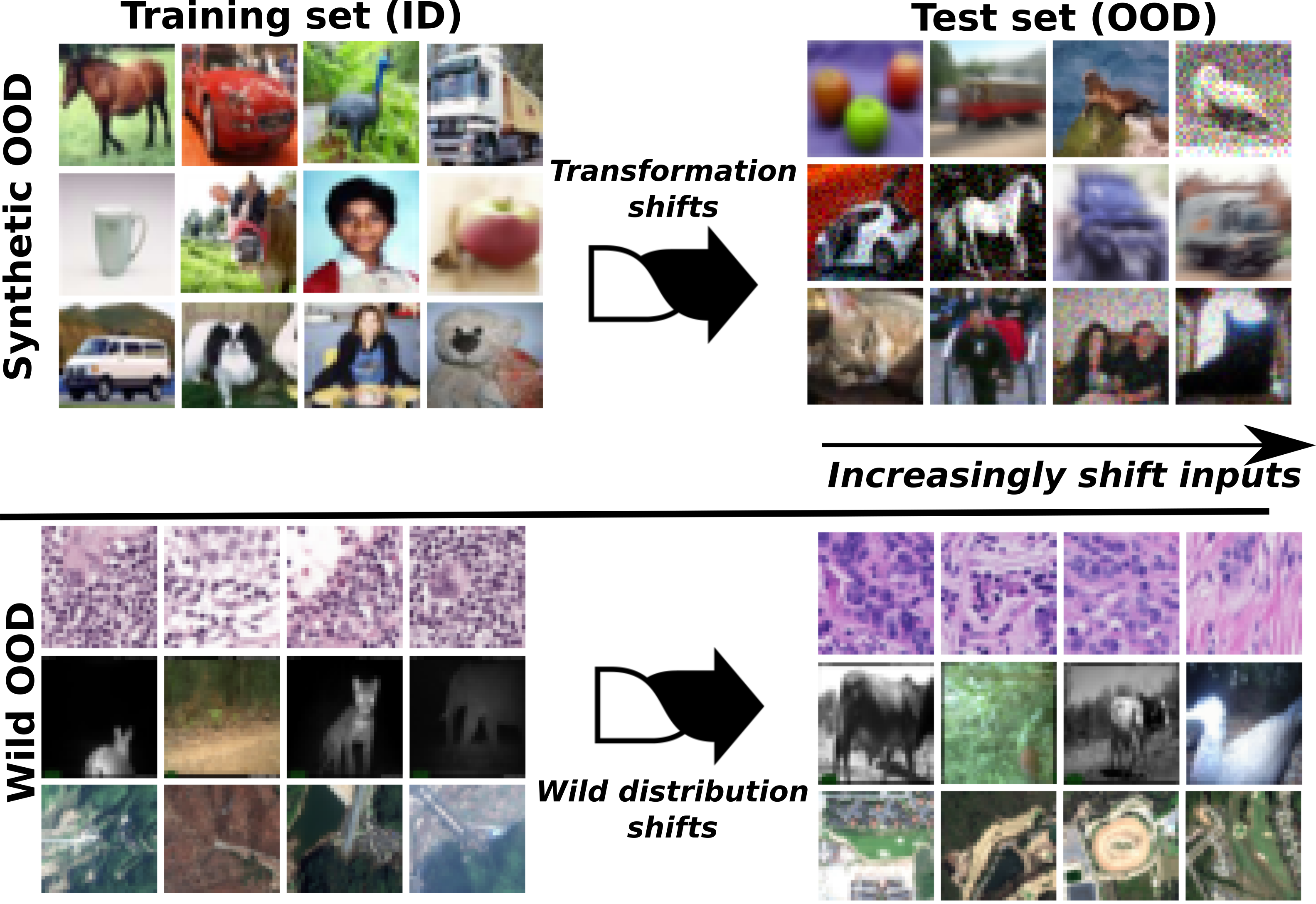}
\caption{Samples from training and augmented validation/test sets for CIFAR10, CIFAR100 and TinyImageNet respectively (synthetic OOD, top 3 rows).  Samples from Camelyon17-Wilds, iWildCam-Wilds and FMoW-Wilds are shown in the bottom 3 rows.}
\label{fig:ood_figures}
\end{figure}

\begin{table*}[!htb]%\setlength\tabcolsep{0.25em}
\centering
\Large
\fontsize{25}{25}\selectfont
\begin{adjustbox}{width=\textwidth}
\begin{tabular}{c|cccc|cccc|cccc}
&\multicolumn{4}{c|}{(a) CIFAR10-C} &\multicolumn{4}{c|}{(b) CIFAR100-C} &\multicolumn{4}{c}{(c) Tiny ImageNet-C}\\
Loss Fn. &Accuracy &ECE &CECE &KSE &Accuracy &ECE &CECE &KSE &Accuracy &ECE &CECE &KSE\\
\hline \hline
CE &77.9$\pm{0.3}$ &14.5$\pm{0.4}$ &3.30$\pm{0.1}$ &14.5$\pm{0.4}$ &52.5$\pm{0.1}$ &10.2$\pm{0.1}$ &0.40$\pm{0.1}$ &7.40$\pm{0.1}$ &25.2$\pm{0.1}$ &15.7$\pm{0.5}$ &0.30$\pm{0.1}$ &15.7$\pm{0.5}$ \\
Focal &77.8$\pm{0.4}$ &9.30$\pm{0.1}$ &2.70$\pm{0.1}$ &9.10$\pm{0.1}$ &52.8$\pm{0.1}$ &8.80$\pm{0.7}$ &0.40$\pm{0.1}$ &8.20$\pm{0.8}$ &24.4$\pm{0.1}$ &13.9$\pm{0.1}$ &0.30$\pm{0.1}$ &13.9$\pm{0.1}$ \\
Inv Focal &77.9$\pm{0.1}$ &16.4$\pm{0.2}$ &3.60$\pm{0.1}$ &16.4$\pm{0.2}$ &52.8$\pm{0.1}$ &13.1$\pm{0.1}$ &0.50$\pm{0.1}$ &11.8$\pm{0.1}$ &26.4$\pm{0.1}$ &16.3$\pm{0.4}$ &0.30$\pm{0.1}$ &16.3$\pm{0.4}$ \\
AvUC &77.8$\pm{0.3}$ &17.8$\pm{0.3}$ &3.80$\pm{0.1}$ &17.5$\pm{0.3}$ &51.2$\pm{0.1}$ &25.4$\pm{0.2}$ &0.60$\pm{0.1}$ &25.0$\pm{0.3}$ &25.9$\pm{0.3}$ &11.3$\pm{0.3}$ &0.30$\pm{0.1}$ &11.2$\pm{0.3}$ \\
Soft AvUC &72.3$\pm{0.1}$ &10.7$\pm{0.1}$ &3.30$\pm{0.1}$ &9.90$\pm{0.2}$ &47.7$\pm{0.1}$ &11.7$\pm{0.1}$ &0.60$\pm{0.1}$ &11.0$\pm{0.1}$ &21.8$\pm{0.5}$ &10.8$\pm{0.2}$ &0.30$\pm{0.1}$ &10.7$\pm{0.2}$ \\
Poly &77.2$\pm{0.2}$ &15.6$\pm{0.4}$ &3.50$\pm{0.1}$ &15.6$\pm{0.4}$ &52.6$\pm{0.1}$ &11.4$\pm{0.3}$ &0.40$\pm{0.1}$ &8.90$\pm{0.3}$ &25.2$\pm{0.2}$ &17.4$\pm{0.1}$ &0.30$\pm{0.1}$ &17.4$\pm{0.1}$ \\
MaxEnt M &77.1$\pm{0.2}$ &\textbf{8.70}$\pm{0.2}$ &2.80$\pm{0.1}$ &\textbf{8.50}$\pm{0.2}$ &52.7$\pm{0.1}$ &\textbf{6.30}$\pm{0.2}$ &\textbf{0.40}$\pm{0.1}$ &\textbf{5.90}$\pm{0.2}$ &22.0$\pm{0.1}$ &10.2$\pm{0.1}$ &0.30$\pm{0.1}$ &10.2$\pm{0.1}$\\
MaxEnt V &76.8$\pm{0.6}$ &8.90$\pm{0.2}$ &\textbf{2.70}$\pm{0.1}$ &8.70$\pm{0.2}$ &51.7$\pm{0.1}$ &8.60$\pm{1.2}$ &0.50$\pm{0.1}$ &8.00$\pm{1.2}$ &21.2$\pm{0.1}$ &\textbf{9.40}$\pm{0.1}$ &\textbf{0.30}$\pm{0.1}$ &\textbf{9.40}$\pm{0.1}$\\
MaxEnt M+V &76.6$\pm{0.5}$ &11.5$\pm{0.5}$ &3.50$\pm{0.1}$ &9.70$\pm{0.5}$ &52.2$\pm{0.1}$ &7.30$\pm{0.5}$ &0.40$\pm{0.1}$ &6.90$\pm{0.4}$ &22.6$\pm{0.1}$ &10.5$\pm{0.1}$ &0.30$\pm{0.1}$ &10.5$\pm{0.1}$ \\
\hline
&\multicolumn{4}{c|}{(d) Camelyon17-Wilds} &\multicolumn{4}{c|}{(e) iWildCam-Wilds} &\multicolumn{4}{c}{(f) FmoW-Wilds}\\
Loss Fn. &Accuracy &ECE &CECE &KSE &Accuracy &ECE &CECE &KSE &Accuracy &ECE &CECE &KSE\\
\hline \hline
CE &81.7$\pm{0.7}$ &15.5$\pm{1.1}$ &16.7$\pm{1.3}$ &15.5$\pm{1.1}$ &52.2$\pm{0.3}$ &30.6$\pm{0.8}$  &0.40$\pm{0.1}$ &30.6$\pm{0.8}$ &35.1$\pm{0.5}$ &39.8$\pm{0.2}$ &1.50$\pm{0.1}$ &39.8$\pm{0.2}$\\
Focal &83.3$\pm{1.6}$ &12.4$\pm{1.7}$ &14.9$\pm{2.0}$ &12.4$\pm{1.7}$ &53.4$\pm{0.4}$ &20.8$\pm{1.2}$  &\textbf{0.30}$\pm{0.1}$ &20.8$\pm{1.2}$ &35.1$\pm{0.5}$ &33.1$\pm{0.6}$ &1.20$\pm{0.1}$ &33.1$\pm{0.6}$\\
Inv Focal &84.3$\pm{2.7}$ &14.2$\pm{2.7}$ &15.0$\pm{2.8}$ &14.2$\pm{2.7}$ &55.9$\pm{0.6}$ &29.9$\pm{0.8}$  &0.40$\pm{0.1}$ &29.9$\pm{0.8}$ &35.2$\pm{0.4}$ &15.7$\pm{2.0}$ &0.70$\pm{0.1}$ &15.7$\pm{2.0}$\\
AvUC &82.6$\pm{0.9}$ &16.0$\pm{0.8}$ &16.5$\pm{1.0}$ &16.0$\pm{0.8}$ &52.6$\pm{0.9}$ &20.2$\pm{1.3}$  &0.40$\pm{0.1}$ &19.7$\pm{1.6}$ &34.9$\pm{0.3}$ &5.80$\pm{1.2}$ &0.50$\pm{0.1}$ &5.80$\pm{1.2}$\\
Soft AvUC &80.0$\pm{3.8}$ &15.6$\pm{2.5}$ &24.0$\pm{1.5}$ &15.7$\pm{2.5}$ &51.1$\pm{0.5}$ &\textbf{18.7}$\pm{2.8}$  &0.50$\pm{0.1}$ &16.3$\pm{4.1}$ &33.1$\pm{0.9}$ &12.7$\pm{0.6}$ &0.80$\pm{0.1}$ &12.7$\pm{0.6}$\\
Poly &80.5$\pm{0.2}$ &17.5$\pm{0.2}$ &18.8$\pm{0.2}$ &17.5$\pm{0.2}$ &54.6$\pm{1.3}$ &27.6$\pm{0.7}$  &0.40$\pm{0.1}$ &27.6$\pm{0.7}$ &35.5$\pm{0.3}$ &15.2$\pm{2.1}$ &0.70$\pm{0.1}$ &15.2$\pm{2.1}$\\
MaxEnt M &82.7$\pm{1.0}$ &12.3$\pm{0.7}$ &14.6$\pm{0.9}$ &12.3$\pm{0.7}$ &53.9$\pm{0.6}$ &20.0$\pm{1.2}$  &0.40$\pm{0.1}$ &20.0$\pm{1.2}$ &34.4$\pm{0.1}$ &5.60$\pm{0.5}$ &0.50$\pm{0.1}$ &5.60$\pm{0.5}$\\
MaxEnt V &83.0$\pm{1.1}$ &11.9$\pm{0.7}$ &13.2$\pm{0.7}$ &11.9$\pm{0.7}$ &50.6$\pm{0.2}$ &25.0$\pm{0.5}$  &0.40$\pm{0.1}$ &25.0$\pm{0.5}$ &33.5$\pm{0.1}$ &\textbf{4.60}$\pm{0.3}$ &\textbf{0.50}$\pm{0.1}$ &\textbf{4.70}$\pm{0.3}$\\
MaxEnt M+V &83.4$\pm{0.9}$ &\textbf{8.30}$\pm{2.0}$ &\textbf{12.2}$\pm{1.9}$ &\textbf{8.30}$\pm{2.0}$ &51.8$\pm{0.5}$ &19.4$\pm{3.6}$  &0.40$\pm{0.1}$ &\textbf{12.2}$\pm{0.5}$ &33.9$\pm{0.3}$ &8.50$\pm{0.3}$ &1.00$\pm{0.2}$ &7.40$\pm{0.3}$\\
\hline
\end{tabular}
\end{adjustbox}
%\vspace{-3mm}
\caption{Test scores (\%) for synthetic (top) and real-world  (bottom) OOD benchmarks computed across different approaches, with our method achieving state-of-the art OOD calibration. $\pm$ indicates the standard errors for 3 random seeds, with the best mean scores highlighted in bold.}
\label{table:results}
\end{table*}

\noindent \textbf{Real-world OOD}: We use the following in-the-wild computer vision datasets with their provided ID training sets and OOD sets for validation and testing. We denote real-world datasets with ``Wilds'' as per \cite{pmlr-v139-koh21a}.
\begin{enumerate}%[noitemsep,nolistsep]
    \item Camelyon17-Wilds \cite{bandi2018detection}: Binary classification task on whether a (32x32) tissue slide contains any malignant/benign tumours.
    \item iWildCam-Wilds \cite{Beery2020TheI2}: Static camera traps deployed across different terrains with radical shifts in camera pose, background and lighting. The task is to identify the species in the photo out of 182 animal classes.
    \item FMoW-Wilds \cite{Fmow}: Satellite images across different functional buildings and terrain from over 200 countries. The task is to detect one out of 62 categories, including a ``false detection'' category.
\end{enumerate}

\subsection{Benchmarking Results}
\subsubsection{Synthetic OOD Evaluation}
We plot the test error, ECE and KSE in Figure \ref{fig:cifar_plots} for each method on the different distribution shifts of CIFAR10/100-C. At level 0 of distribution shift (ID test set) we find that most methods are relatively well calibrated with $\le 5\%$ ECE. However, they tend to become miscalibrated with higher test errors as the distributions increasingly shift  away from the training set, with the poorest calibration coming from InvFocal and AvUC loss on both sets of CIFAR. On the other hand, Focal and Soft-AvUC loss perform relatively well OOD, with the best performance coming from our method. We note that the authors' original results for Soft-AvUC did not beat the Focal loss baseline, whereas in our experiments we choose the hyperparameters ($T, \kappa$) to the best of our ability, resulting in better performance than Focal loss OOD. %For this CIFAR example, the mean constraint form of our method (in blue) performs best.

\begin{figure}[!tb]
\includegraphics[width=\columnwidth]{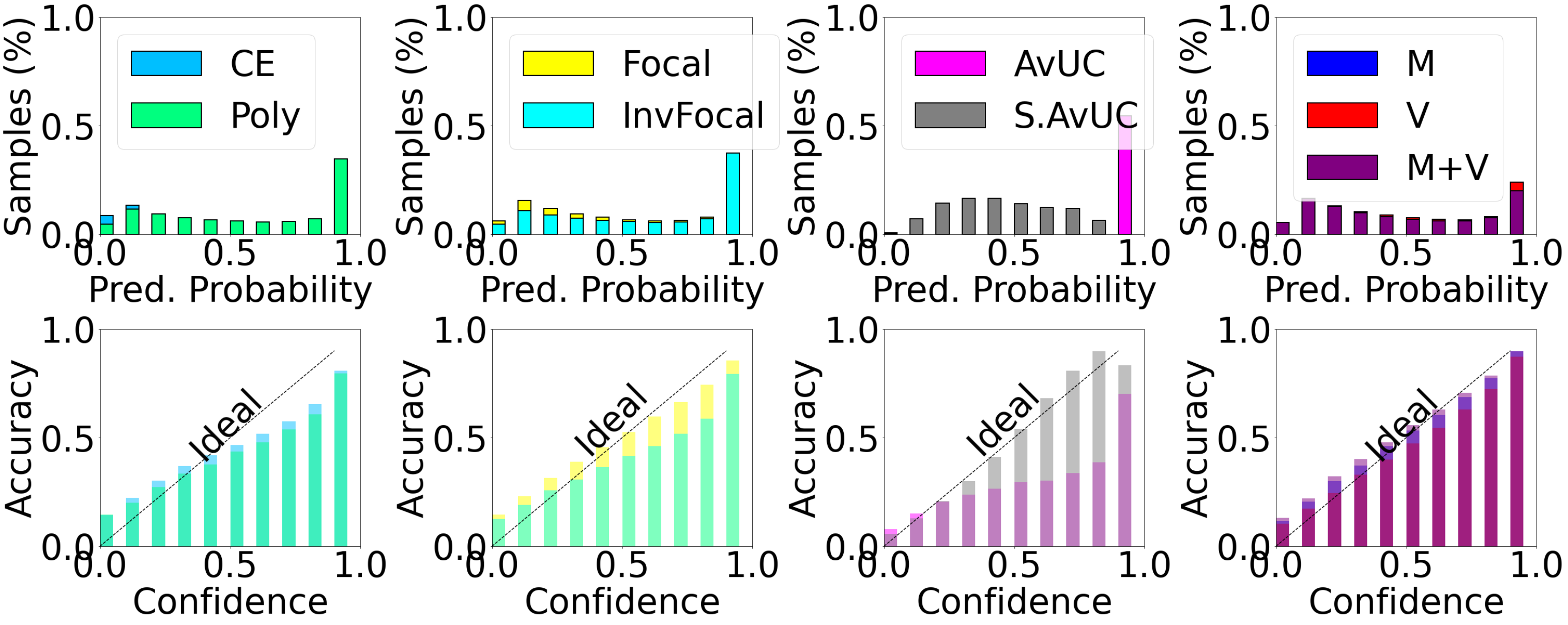}
%\vspace{-6mm}
\caption{Bin-strength densities (top) and reliability diagrams (bottom) computed using $B=10$ bins for different loss functions, evaluated on CIFAR100-C. MaxEnt Loss delivers a more uniform spread of probability densities and a reliability bar plot that better matches the ideal diagonal.}
\label{fig:probabilty_densities}
\end{figure}

We compare the bin-strength and reliability diagrams \cite{NiculescuMizil2005PredictingGP} for the different loss functions averaged across distribution shifts of CIFAR100-C in \cref{fig:probabilty_densities}. The bin-strengths plots show that the predictions of most methods remain concentrated in high-confidence bins, which suggests that these loss functions produce ``peaky'' distributions and over-confidence. In contrast, all three forms of our method mean (M), variance (V), mean plus variance (M+V) provide a better spread of predictions across bins, having significantly ``softer'' probabilities. From the reliability diagrams, the bars of most methods lie below the ideal diagonal, indicating that their predictions are over-confident and miscalibrated. On the other hand, our method produces bars that follow the diagonal more closely, demonstrating better calibration performance.

We further show the OOD results on the CIFAR10/100-C and TinyImageNet-C test sets, namely the accuracy, ECE, CECE (both computed with 15 bins) and KSE in \cref{table:results}. For the experiments shown in these tables, we use one-hot labels during training and report their performance without temperature scaling. We report the mean scores and the standard error (standard deviation divided by the square root of random seeds). When evaluated ID, all methods produce similar accuracies and are roughly within 95\%, 75\% and 50\% for the original test sets of CIFAR and TinyImageNet respectively. 

\begin{table*}[!htb]
\fontsize{20}{20}\selectfont
\center
\begin{adjustbox}{width=\textwidth}
\begin{tabular}{|l|ccc|ccc|ccc|}
\hline
& MaxEnt &MaxEnt  &MaxEnt &MaxEnt &MaxEnt &MaxEnt &MaxEnt &MaxEnt &MaxEnt \\
Dataset & Mean & Var &Mult & Mean w/ TS & Var w/ TS &Mult w/ TS & Mean w/ LS & Var w/ LS &Mult w/ LS \\
\hline
CIFAR10-C &8.70$\pm{0.2}$ &8.90$\pm{0.2}$ &11.5$\pm{0.5}$ &$\downarrow$6.90$\pm{0.1}$ &$\downarrow$\textbf{6.70}$\pm{0.2}$ &$\downarrow$9.30$\pm{0.5}$ &9.60$\pm{0.7}$ &9.80$\pm{0.9}$ &11.6$\pm{0.1}$\\
CIFAR100-C &\textbf{6.30}$\pm{0.2}$ &8.60$\pm{1.2}$ &7.30$\pm{0.5}$ &11.1$\pm{0.1}$ &$\downarrow$7.60$\pm{0.2}$ &12.3$\pm{0.4}$ &15.4$\pm{2.5}$ &13.0$\pm{2.4}$ &16.2$\pm{2.8}$\\
TinyImageNet-C &10.2$\pm{0.1}$ &9.40$\pm{0.1}$ &10.5$\pm{0.1}$ &10.4$\pm{0.1}$ &10.4$\pm{0.1}$ &$\downarrow$9.60$\pm{0.1}$ &$\downarrow$\textbf{9.00}$\pm{1.0}$ &9.50$\pm{0.1}$ &$\downarrow$9.60$\pm{0.8}$\\
\hline
Camelyon17-Wilds &12.3$\pm{0.7}$ &11.9$\pm{0.7}$ &8.30$\pm{2.0}$ &$\downarrow$7.40$\pm{1.4}$ &$\downarrow$7.40$\pm{0.5}$ &$\downarrow$\textbf{4.60}$\pm{1.1}$ &$\downarrow$6.50$\pm{0.1}$ &$\downarrow$7.00$\pm{1.0}$ &9.40$\pm{0.4}$\\
iWildCam-Wilds &20.0$\pm{1.2}$ &25.0$\pm{0.5}$ &19.4$\pm{3.6}$ &$\downarrow$8.20$\pm{0.8}$ &$\downarrow$\textbf{7.40}$\pm{1.0}$ &$\downarrow$11.8$\pm{2.0}$ &$\downarrow$7.40$\pm{1.1}$ &$\downarrow$9.20$\pm{0.4}$ &$\downarrow$18.5$\pm{0.1}$\\
FmoW-Wilds &5.60$\pm{0.5}$ &4.60$\pm{0.3}$ &8.50$\pm{0.3}$ &$\downarrow$\textbf{3.50}$\pm{0.4}$ &$\downarrow$4.40$\pm{0.3}$ &$\downarrow$4.00$\pm{0.9}$ &6.20$\pm{0.9}$ &4.60$\pm{0.3}$ &$\downarrow$7.60$\pm{2.3}$\\
\hline
\end{tabular}
\end{adjustbox}
%\vspace{-3mm}
\caption{ECE (\%) scores showcasing the effects of temperature scaling (TS) and label smoothing (LS) for the different OOD datasets. $\downarrow$ indicates improvements over the baseline and $\pm$ shows the standard errors with the best scores highlighted in bold.}
\label{table:ad-hoc}
\end{table*}

\subsubsection{Real-world OOD Evaluation}
We further evaluate our method on real-world datasets, where distribution shifts  occur naturally in-the-wild. Contrary to the synthetic datasets, the prior distribution for the Wilds datasets can be non-uniform, which might result in some form of bias for certain classes. Regardless, this does not negatively impact the performance of our method, as shown in \cref{table:results}. In general, we observe similar results as for synthetic OOD. Firstly, most loss functions produce relatively similar recognition accuracies across all datasets, regardless of synthetic or wild OOD. There are no significant drops in test set recognition when models are trained using our method. Secondly, models trained with MaxEnt loss are competitive and generally achieve state-of-the art performance alongside other baselines in terms of the various calibration metrics.

Importantly, MaxEnt loss is able to consistently deliver well calibrated models, even without the use of ad-hoc calibration techniques such as temperature scaling. We notice that the three forms of MaxEnt loss produce roughly similar performance, which is unsurprising since the constraints come from the same training set. However, as illustrated in \cref{fig:simple_maxent}, these entropies are maximized subject to the given constraints, which can yield different results after optimization. This helps to explain why combining the mean and variance constraints does not necessarily lead to better calibration. In the Appendix, we report similar results when evaluating our method with different calibration metrics on both synthetic and wild OOD. %We are unable to prescribe any particular method that will lead to the best OOD results and hypothesize that the final performance of each loss function would still depend on the distribution and nature of the given dataset. 

\subsection{Pre- and Post-hoc Calibration}
In this section, we discuss how MaxEnt loss complements commonly used ad-hoc techniques such as label smoothing and temperature scaling.

\noindent \textbf{Pre-hoc Calibration:} Label smoothing artificially softens the target distribution and seeks to encourage high entropy predictions \cite{Mller2019WhenDL}. Formally, the smoothed vector $s_i$ is obtained after \textit{uniformly} redistributing the probabilities of the correct class to other classes by a smoothing factor $\alpha$: $s_i = (1 - \alpha) y_k + \frac{\alpha}{K}$. %As a pre-hoc calibration technique, label smoothing is highly dependent on the hyper-parameter $\alpha$: If $\alpha=0$, the original one-hot vector $y_k$ is kept; when $\alpha=1$, the uniform distribution is obtained. 

In contrast, our method performs smoothing \textit{non-uniformly} with the help of constraints derived from the training distribution. In Appendix C.3, we can see that label smoothing only applies a linear shift in the bin-strength densities, whereas MaxEnt loss delivers different distributions with different global constraints. When label smoothing is applied together with our method, we are able to obtain models with even ''flatter'' bin-strength densities.  We show the performance of our method with and without label smoothing in the right column of \cref{table:ad-hoc}, using $\alpha=0.01$. Label smoothing is only able to improve the ECE of our method in some cases. If the model is already well-calibrated, label smoothing can negatively affect calibration. 

\noindent \textbf{Post-hoc Calibration:} For post-hoc calibration, we choose the standard temperature scaling technique, which linearly scales the classifier's output logits with a scalar $T>0$. We follow the recommendations of \citet{mukhoti2020calibrating} and perform grid-search over a typical range of temperature values [1.25, 1.50, 1.75, 2.00], picking the optimal temperature that minimizes the negative log-likelihood (NLL) \cite{hastie01statisticallearning} of the validation set. We highlight that temperature scaling does not affect accuracy and is only helpful for model calibration under i.i.d.\ assumptions of the test set \cite{Ovadia2019CanYT}. In the case where the validation set is ID and not OOD shifted, temperature scaling may not be useful for improving calibration \cite{karandikar2021soft}. As compared to label smoothing, our method exhibits the best calibration performance after temperature scaling, with significant improvements in ECE. This is particularly encouraging, because our method complements temperature scaling and does not restrict users with regards to post-hoc calibration.

\begin{figure}[!b]
\small
\includegraphics[width=\columnwidth]{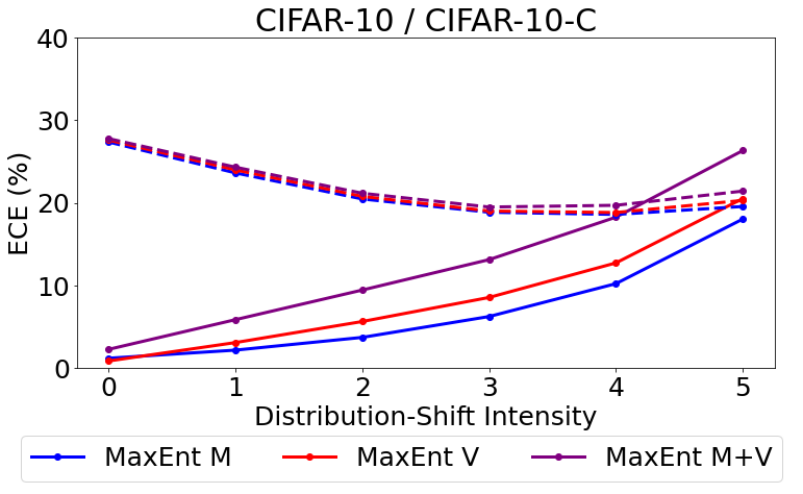}
\caption{With only global constraints (dotted), predictions tend to be underconfident. Better calibration can be achieved by combining both global and local constraints (solid).}
\label{local_constraints}
\end{figure}

\subsection{Effects of Local Constraints}
\cref{local_constraints} shows the empirical calibration error curves on CIFAR10-C, where we only consider the global constraints for each form of MaxEnt loss versus when the local constraints are included. 

Firstly, without local constraints all three forms of our methods tend to be miscalibrated and underconfident for the lower intensity shifts. Secondly, all three global forms tend towards the same calibration performance across all shifts. With the inclusion of local constraints, all three forms of our method become better calibrated on lower intensity shifts, with the best calibration performance achieved by the mean constraint form.

\subsection{Ordering of Feature Norms}
\begin{figure}[!tb]
\small
\includegraphics[width=\columnwidth]{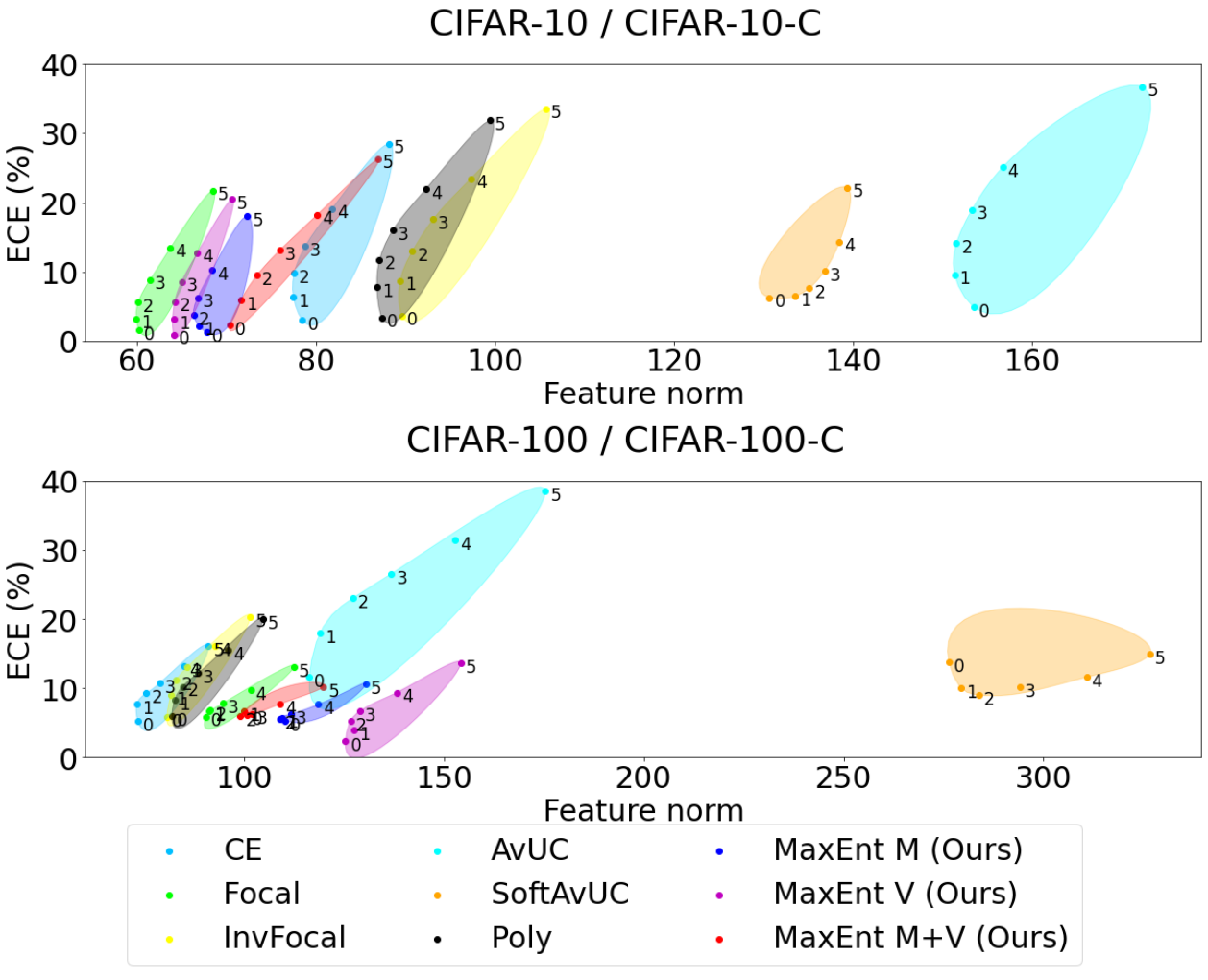}
\caption{ECE vs.\ L2 norm of features from the penultimate layer of ResNet-18 trained using different methods, compared across CIFAR-C. Our method produces lower errors with smaller norms.}
\label{fig:error_norm_ood} 
\end{figure}

We further analyze the performance of each loss function and show the ECE scatter plot as a function of the L2 norm of the learnt features in \cref{fig:error_norm_ood}. Specifically, each numbered point represents the average ECE per feature norm for the six intensity shifts extracted from the logits of the penultimate layer. We observe a correlation between ECE and feature norms, where lower feature norms tend to deliver smaller calibration errors. Apart from AvUC and SoftAvUC, most methods tend to support the findings from \cite{pmlr-v70-guo17a} that there is a strong relationship between miscalibration and overfitting. Our method generally produces small clusters with low calibration errors, which suggests that constraints provide robustness against distribution shifts and overfitting.

\section{Limitations and Future Work}
\textbf{Choice of Constraints:} In the current setup, we attribute the improvements in calibration to the given constraints. This assumes that the ratio/prior distribution $P(\mathcal{Y})$ observed during training is aligned with the test distribution. For test distributions that vary greatly from the training set, we believe that our method (along with many others) would not fare well. Since it is difficult to know the true distribution of the task (e.g.\ during model deployment), further approximations may be needed.

\noindent \textbf{Unique Lagrange Multipliers:} Our proposed framework approximates the Lagrange multipliers for each form of our method. Currently, we use a single Lagrange multiplier shared universally across both the global and local constraints. It may be possible to introduce unique Lagrange multipliers to control the trade-off between global and local constraints. However, this is non-trivial since it would require careful tuning and selection of hyperparameters, which is largely dependant on a separate validation set. %We further show in the Appendix, that these Lagrange multipliers are non-binding and satisfy the subjected conditions.

\section{Conclusion}
We presented MaxEnt loss, a novel loss function with multiple forms for calibrating deep neural networks across both synthetic and in-the-wild OOD computer vision datasets. We also showed the relationship between Focal loss and the Principle of Maximum Entropy. MaxEnt Loss achieves state-of-the art calibration with no significant increase in computation costs and requires only a few additional lines of code. Predictive uncertainty typically worsens under increasing dataset shift, whereas MaxEnt loss remains robust without any additional ad-hoc calibration. Furthermore, MaxEnt loss complements other ad-hoc calibration methods such as temperature scaling and label smoothing.

\section{Acknowledgements}
This research is supported in part by the Ministry of Education, Singapore, under its MOE AcRF TIER 3 Grant (MOE-MOET32022-0001). The authors would like to thank Tan Yingkiat and Bai Yunwei for their helpful suggestions and feedback.

\bibliography{aaai24}
\clearpage
\medskip

\appendix
\onecolumn

%Structure:
%A.) Mathematical Derviations
%B.) Datasets and Training Details
%C.) Additional Metrics and Results

\section{Mathematical Derivations}
\label{proofs}
\subsection{Relation between Focal and MaxEnt Loss for Entropy Regularised KL Divergence}
Using the following proof inspired by \cite{mukhoti2020calibrating}, by including additional constraints, then MaxEnt Loss is given in the form of  Entropy Regularized KL divergence subjected to constraints:
\begin{fleqn}[\parindent]
\begin{align*}
    \mathcal{L_F} &=  - \sum^{K}_{k=1} (1 - P_i(y_k|x_i) )^\gamma y_k \log P_i(y_k|x_i)  \\
    &\ge - \sum^{K}_{k=1} (1 - \gamma P_i(y_k|x_i) ) y_k \log P_i(y_k|x_i) \quad \text{By Bernoulli's Inequality}\\
    &= - \sum^{K}_{k=1} y_k \log P_i(y_k|x_i) - \gamma |\sum^{K}_{k=1} y_k P_i(y_k|x_i) \log P_i(y_k|x_i)  | \\
    &\ge - \sum^{K}_{k=1} y_k \log P_i(y_k|x_i) - \gamma \max y_k \sum^{K}_{k=1} |P_i(y_k|x_i) \log P_i(y_k|x_i)| \\
    &\ge - \sum^{K}_{k=1} y_k \log P_i(y_k|x_i) + \gamma \sum^{K}_{k=1} P_i(y_k|x_i) \log P_i(y_k|x_i) \\
    \mathcal{L_{ME}} &\ge \underbrace{- \sum^{K}_{k=1} y_k \log P_i(y_k|x_i)}_{\text{CE loss}} + \underbrace{\gamma \sum^{K}_{k=1} P_i(y_k|x_i) \log P_i(y_k|x_i)}_{\text{Entropy}} + \underbrace{\sum^M_{n=1} + \lambda_n \left( \sum^K_{k=1} \mathcal{Y} P_i(y_k|x_i) - c_n \right)}_{\text{Constraints}}
    \end{align*}
    \label{relation}
\end{fleqn}
\subsection{Computing the Expected Mean and Variance}
Consider the following example from the training set of CIFAR10 and it's prior distribution $P(\mathcal{Y})$ \emph{after} randomly splitting out samples from each of the ten classes for validation, where:
\begin{equation}
\centering
\begin{split}
&\text{For each class:} \quad \mathcal{Y} = \bigr[ 0, 1, 2, 3, 4, 5, 6, 7, 8, 9 \bigr] \\
&P(\mathcal{Y}) = \bigr[0.0996, 0.1002, 0.0987, 0.1009, 0.0995, 0.0996, 0.1006, 0.0997, 0.1005,0.1006\bigr]
\end{split}
\end{equation}
With the above given prior distribution, the expected global average of the given target distribution can therefore be computed as:
\begin{equation}
\begin{split}
\E[\mathcal{Y}] = \mu_G \quad \\
= \sum_k \mathcal{Y} P(\mathcal{Y})
\label{mean_constraint}
\end{split}
\end{equation}
Which, for this particular CIFAR10 example is calculated as $\mu_G = 4.5082$. Subsequently, the expected variance of the given target distribution also be computed accordingly by reusing the value of the expected average and using the handy relationship $\E[\mathcal{Y}^2] - \E[\mathcal{Y}]^2$:
\begin{equation}
\begin{split}
\E[\mathcal{Y}^2] - \E[\mathcal{Y}]^2 = \sigma^2 \\
= \sum_k \mathcal{Y}^2 P(\mathcal{Y}) - \mu^2
\end{split}
\end{equation}
Which gives $\sigma^2 = 8.2572$. Using these computed constraints, we can further compute the individual Lagrange multipliers in the following subsections for each of the constrained MaxEnt methods proposed in Equations \ref{expo_final}, \ref{gaussian_final} and \ref{multiple_final}. \label{mathematical_derivation}
\pagebreak

\subsection{Proofs for Lagrange Multipliers}
Here we provide the mathematical formulations for the Maximum Entropy objective functions presented in the main text. With some rigorous mathematical derivations, the Lagrange multipliers can be solved simply using a root-finder (e.g.\ Newton Raphson's) which is used in our implementation and also made available in other open-sourced codes. 
\subsubsection{Principle of Maximum Entropy: Mean Constraint}
Suppose the expected average $\mu$ of the target distribution from \cref{mean_constraint} is known, then the following constraints can be added to the unconstrained form of the MaxEnt method in \cref{max_ent_method}. The mean constraint is also commonly referred to as the first order momentum constraint:
\begin{fleqn}[\parindent]
    \begin{align*}
    \centering
       &\mathcal{L} = - \sum_k P_i(y_k|x) \log P_i(y_k|x) - \lambda_G (\sum_k \mathcal{Y} P_i(y_k|x) - \mu_G) - \lambda_L (\sum_k \mathcal{Y} P_i(y_k|x) - \mu_Lk) \\
       &\textit{Expand the Lagrangian, take its derivative and set to zero}\\
       &\frac{\partial \mathcal{L}}{\partial P_i(y_k|x)} = \log P_i(y_k|x) + 1 + (\lambda_G + \lambda_L) \mathcal{Y} = 0 \tag{$A$}\\
       &\textit{Subject to the following inequality constraints}\\
       &\frac{\partial \mathcal{L}}{\partial \lambda_G} = \sum_k \mathcal{Y} P_i(y_k|x) - \mu_G \le 0 \tag{$B$}\\
       &\frac{\partial \mathcal{L}}{\partial \lambda_L} = \sum_k \mathcal{Y} P_i(y_k|x) - \mu_L \le 0 \tag{$C$}\\
       &\textit{Rearrange Equation A and let $(\lambda_G + \lambda_L) =\lambda_\mu$ to obtain general solution}\\
       &P_i(y_k|x) = e^{- 1 - (\lambda_G + \lambda_L) \mathcal{Y}} = e^{- 1 - \lambda_\mu \mathcal{Y}} \tag{$D$}\\ 
       &\textit{Satisfy both global and local constraints by combining such that $\mu = \frac{\mu_G + \mu_L}{2}$}\\
       &\sum_k \mathcal{Y} P_i(y_k|x) = \sum_k e^{- 1 - \lambda_\mu \mathcal{Y}}\mathcal{Y} = \mu\\
       &\sum_k e^{- \lambda_\mu \mathcal{Y}} \mathcal{Y} = \frac{\mu}{e^{-1}} \\
       &\textit{Approximate $\lambda_\mu$ using Newton's method with the following helper function and it's derivative } \\
       &g(\lambda_\mu) = \sum_k e^{-\lambda_\mu\mathcal{Y}} \mathcal{Y}\\
       &g'(\lambda_\mu) = \sum_k e^{-\lambda_\mu \mathcal{Y}} \mathcal{Y}^2 + \sum_k e^{-\lambda_\mu \mathcal{Y}} \\
    \end{align*}
\end{fleqn}
\clearpage
\subsubsection{Principle of Maximum Entropy: Variance Constraint}
Suppose the expected variance $\sigma^2$ of the target distribution is known, then the following variance constraint is added to Equation \ref{max_ent_method}. As mentioned in the main paper, we do not consider any knowledge of the mean constraint $\mu$ in this form. The variance constraint is also commonly referred to as the second order momentum constraint and is given by: 
\begin{align*}
   &\mathcal{L} = - \sum_k P_i(y_k|x) \log P_i(y_k|x) - \lambda_G (\sum_k \mathcal{Y}^2 P_i(y_k|x) - \sigma^2_G) - \lambda_L (\sum_k \mathcal{Y}^2P_i(y_k|x) - \sigma^2_Lk) \\
   &\textit{Expand the Lagrangian, take its derivative and set to zero}\\
   &\frac{\partial \mathcal{L}}{\partial P_i(y_k|x)} = \log P_i(y_k|x) + 1 + (\lambda_G + \lambda_L) \mathcal{Y}^2 = 0 \tag{$A$}\\
   &\textit{subject to the following inequality constraints}\\
   &\frac{\partial \mathcal{L}}{\partial \lambda_G} = \sum_k \mathcal{Y}^2 P(y_k|x_i) - \sigma^2_G \le 0 \tag{$B$}\\
   &\frac{\partial \mathcal{L}}{\partial \lambda_L} = \sum_k \mathcal{Y}^2 P(y_k|x_i) - \sigma^2_L \le 0 \tag{$C$}\\
   &\textit{Rearrange Equation A and let $(\lambda_G + \lambda_L) = \lambda_{\sigma^2}$ obtain general solution}\\
   &P_i(y_k|x) = e^{- 1 - \lambda_{\sigma^2} \mathcal{Y}^2} \tag{$D$}\\ 
   &\textit{Satisfy both global and local constraints by combining such that $\sigma^2 = \frac{\sigma^2_G + \sigma^2_L}{2}$}\\
   &\sum_k P_i(y_k|x) \mathcal{Y}^2 = \sum_k e^{- 1 - \lambda_{\sigma^2} \mathcal{Y}^2} \mathcal{Y}^2 = \sigma^2 \\
   &\sum_k e^{- \lambda_{\sigma^2} \mathcal{Y}^2} \mathcal{Y}^2 = \frac{\sigma^2}{e^{-1}} \\
   &\textit{Solve for $\lambda_2$ using Newton's method with the following helper function and it's derivative} \\
   &g(\lambda_{\sigma^2}) = \sum_k e^{-\lambda_{\sigma^2} \mathcal{Y}^2} \mathcal{Y}^2\\
   &g'(\lambda_{\sigma^2}) = \sum_k e^{-\lambda_{\sigma^2} \mathcal{Y}^2}  \mathcal{Y}^4 + \sum_k e^{-\lambda_2 \mathcal{Y}^2} \\
\end{align*}
\subsubsection{Principle of Maximum Entropy: Multiple Constraints}
Finally, suppose both the expected average $\mu$ and variance $\sigma^2$ constraints are known. In this form we consider both constraints jointly with the variance defined as $\mathbb{E}[{\mathcal{Y} - \mu}^2]$. The constraints are then added to the loss function, where the Lagrange multipliers have to be solved simultaneously. Consider the following multiple constraint objective:
\begin{align}
\begin{split}
   \mathcal{L} = - \sum_k P_i(y_k|x) \log (P_i(y_k|x)) - \lambda_{1_G} (\sum_k P_i(y_k|x)\mathcal{Y} - \mu_G) - \lambda_{1_L} (\sum_k P_i(y_k|x)\mathcal{Y} - \mu_{Lk}) \\ 
   - \lambda_{2_G} (\sum_k P_i(y_k|x)(\mathcal{Y} - \mu)^2 - \sigma^2_G) - \lambda_{2_L} (\sum_k P_i(y_k|x)(\mathcal{Y} - \mu)^2 - \sigma^2_{Lk}) \\
\end{split}
\end{align}
\begin{align*}
   &\textit{Expand the Lagrangian above, take its derivative and set to zero}\\
   &\frac{\partial \mathcal{L}}{\partial P_i(y_k|x)} = \log (P_i(y_k|x)) + 1 + (\lambda_{1_G} + \lambda_{1_L}) \mathcal{Y} + (\lambda_{2_G} + \lambda_{2_L}) (\mathcal{Y} - \mu)^2 = 0 \tag{$A$}\\
   &\frac{\partial \mathcal{L}}{\partial P_i(y_k|x)} = \log (P_i(y_k|x)) + 1 + \lambda_1 \mathcal{Y} + \lambda_2 (\mathcal{Y} - \mu)^2 = 0 \tag{$A$}\\
   &\textit{Subject to the following inequality constraints}\\
   &\frac{\partial \mathcal{L}}{\partial \lambda_{1G}} = \sum_k \mathcal{Y} P_i(y_k|x) - \mu_G \le 0 \tag{$B$}\\
   &\frac{\partial \mathcal{L}}{\partial \lambda_{1L}} = \sum_k \mathcal{Y} P_i(y_k|x) - \mu_L \le 0 \tag{$C$}\\
   &\frac{\partial \mathcal{L}}{\partial \lambda_{2G}} = \sum_k (\mathcal{Y} - \mu)^2 P_i(y_k|x) - \sigma^2_G \le 0 \tag{$C$}\\
   &\frac{\partial \mathcal{L}}{\partial \lambda_{2L}} = \sum_k (\mathcal{Y} - \mu)^2 P_i(y_k|x) - \sigma^2_L \le 0 \tag{$C$}\\
   &\textit{Rearrange Equation A let $(\lambda_{1G} + \lambda_{1L}) = \lambda_1$ and $(\lambda_{2G} + \lambda_{2L}) = \lambda_2$ to obtain general solution}\\   
   &P_i(y_k|x) = e^{- 1 -\lambda_1 \mathcal{Y} - \lambda_2 (\mathcal{Y} - \mu)^2} \tag{$D$}\\ 
   &\textit{Satisfy both global and local constraints by combining such that $\mu = \frac{\mu_G + \mu_L}{2}$ and $\sigma^2 = \frac{\sigma^2_G + \sigma^2_L}{2}$ }\\
   &\sum_k P_i(y_k|x)\mathcal{Y} = \sum_k e^{- 1 -\lambda_1 \mathcal{Y} - \lambda_2 (\mathcal{Y} - \mu)^2} \mathcal{Y} = \mu \\
   &\sum_k P_i(y_k|x)(\mathcal{Y} - \mu)^2 = \sum_k e^{- 1 -\lambda_1 \mathcal{Y} - \lambda_2 (\mathcal{Y} - \mu)^2} (\mathcal{Y} - \mu)^2 = \sigma^2 \\
   &\sum_k e^{-\lambda_1 \mathcal{Y} - \lambda_2 (\mathcal{Y} - \mu)^2} \mathcal{Y} = \frac{\mu_\theta}{e^{-1}} \\
   &\sum_k e^{-\lambda_1 \mathcal{Y} - \lambda_2 (\mathcal{Y} - \mu)^2} (\mathcal{Y} - \mu)^2 = \frac{\sigma^2}{e^{-1}} \\
   &\textit{Solve for $\lambda_1 \text{and} \lambda_2$ using Newton's method with the following helper functions and their Jacobian matrix $J(\lambda_n)$} \\
   &\lambda_{n+1} = \lambda_n - J ^{-1}(\lambda_n) g(\lambda_n)\\
   &\textit{Where in matrix form, is written as} \\
   &\begin{bmatrix}
    \lambda_{1+1}\\
    \lambda_{2+1}
    \end{bmatrix} =
    \begin{bmatrix}
    \lambda_{1}\\
    \lambda_{2}
    \end{bmatrix}
    \begin{bmatrix}
    \frac{\partial g_1 }{ \partial \lambda_1} & \frac{\partial g_1 }{ \partial \lambda_2}\\
    \frac{\partial g_2 }{ \partial \lambda_1} & \frac{\partial g_2 }{ \partial \lambda_2}
    \end{bmatrix}^{-1}
    \begin{bmatrix}
    g_1\\
    g_2
    \end{bmatrix} \\
   &\textit{Specifically, the elements for the Jacobian contains the partial derivatives of each of the helper functions, where:} \\
   &g_1(\lambda_1, \lambda_2) = \sum_k e^{-\lambda_1 \mathcal{Y} - \lambda_2 (\mathcal{Y} - \mu)^2} \mathcal{Y}\\
   &g_2(\lambda_1, \lambda_2) = \sum_k e^{-\lambda_1 \mathcal{Y} - \lambda_2 (\mathcal{Y} - \mu)^2} (\mathcal{Y} - \mu)^2\\
   & \frac{\partial g_1(\lambda_1, \lambda_2) }{ \partial \lambda_1} = \sum_k e^{-\lambda_1 \mathcal{Y} - \lambda_2 (\mathcal{Y} - \mu)^2} \mathcal{Y}^2 + \sum_k e^{-\lambda_1 \mathcal{Y} - \lambda_2 (\mathcal{Y} - \mu)^2} \\
   & \frac{\partial g_1(\lambda_1, \lambda_2) }{ \partial \lambda_2} =  \sum_k e^{-\lambda_1 \mathcal{Y} - \lambda_2 (\mathcal{Y} - \mu)^2} \mathcal{Y} (\mathcal{Y} - \mu)^2 + \sum_k e^{-\lambda_1 \mathcal{Y} - \lambda_2 (\mathcal{Y} - \mu)^2} \\
   & \frac{\partial g_2(\lambda_1, \lambda_2) }{ \partial \lambda_1} =  \sum_k e^{-\lambda_1 \mathcal{Y} - \lambda_2 (\mathcal{Y} - \mu)^2} \mathcal{Y} (\mathcal{Y} - \mu)^2 + \sum_k e^{-\lambda_1 \mathcal{Y} - \lambda_2 (\mathcal{Y} - \mu)^2} \\
   & \frac{\partial g_2(\lambda_1, \lambda_2) }{ \partial \lambda_2} =  \sum_k e^{-\lambda_1 \mathcal{Y} - \lambda_2 (\mathcal{Y} - \mu)^2} (\mathcal{Y} - \mu)^4 + \sum_k e^{-\lambda_1 \mathcal{Y} - \lambda_2 (\mathcal{Y} - \mu)^2} \\
   %&\textit{Coincidentally, elements $\frac{\partial g_1(\lambda_1, \lambda_2) }{ \partial \lambda_2} $ $\frac{\partial g_2(\lambda_1, \lambda_2) }{ \partial \lambda_1}$ are equivalent and the resultant Jacobian is a symmetrical matrix.}
\end{align*}
\clearpage
\subsection{Sanity Check (Mean Constraint)}
We can easily check the correctness of the implementation, by simply substituting the Lagrange multipliers into the constraints. For the CIFAR10 example we use the mean constraint with the global and local expected values $\mu_G=4.5082$ and for class 1 $\mu_{L1}=1$. Following these constraints, this leads to $\mu = \frac{\mu_G + \mu_L}{2} = 2.7541$ where the Newton's method returns the following value for $\lambda_\mu=0.3294$. Substituting these values into the general solution obtained previously.
\begin{align*}
    &\sum_k e^{- 1 -\lambda_\mu \mathcal{Y}} \mathcal{Y} \le \mu\\
    &\sum_k e^{- 1 -0.3294 \mathcal{Y}} \mathcal{Y} \le 2.7541 \tag{$\underline{True}$}\\
    &\textit{The equation above is \underline{True} since the general solution gives:}\\
    &\sum_k e^{- 1 -0.3294 \mathcal{Y}} \mathcal{Y} = 2.7489 \le 2.7541
\end{align*}

\section{OOD Datasets and Training Details}
\label{dataset_details}
Here, we provide the dataset descriptions used in the main paper with an illustration of examples in \cref{fig:ood_figures}, along with additional training details such as the hyperparameters used in the main paper. For synthetic OOD, the corrupted forms of CIFAR-C and TinyImageNet-C contain a total of 19 different image transformations. Specifically the following image augmentations

\begin{table}[!htb]
    \centering
    \begin{tabular}{c|c|}
        &'gaussian noise', 'shot noise', 'impulse noise', 'defocus blur'\\ 
        &\qquad 'motion blur', 'zoom blur', 'snow', 'frost', 'fog', 'brightness', \\
        &\qquad 'contrast', 'elastic transform', 'pixelate', 'jpeg compression', \\
        &\qquad 'glass blur', 'speckle noise', 'gaussian blur', 'spatter', 'saturate'\\
    \end{tabular}
    \caption{List of image transformations for CIFAR and TinyImageNet}
    \label{tab:my_label}
\end{table}
\noindent were applied on the CIFAR test set and TinyImageNet validation set. For the Wilds datasets, distribution shifts can come from variations in different hospital collection processes (Camelyon17), different locations of camera traps (iWildCam) and disparities between regions and continents (FMoW). 

For our experiments on synthetic OOD, we make use ResNet-18 for CIFAR10/100 and ResNet-50 \cite{He2016DeepRL} for TinyImageNet with SGD optimizer momentum of 0.9 and learning rate at 0.1 with Cosine Annealing scheduler. For Wild OOD, we use ResNet-18 and DenseNet-121 \cite{Huang2017DenselyCC} with Adam optimizer \cite{Kingma2015AdamAM} with a constant learning rate fixed at 2.5e-4. We train 100 epochs with a batch size of 512 for CIFAR datasets and 50 epochs with a batch size of 256 for wild OOD datasets. 

For comparison against other loss functions, we closely reference the implementations provided by the respective authors, and show their performance on both synthetic and Wild OOD. We choose CE, Focal ($\gamma$=1), Inverse Focal ($\gamma$=2), AvUC ($\beta$=3), Soft AvUC ($T$=0.01, $\kappa$=0.1) ($\gamma$=1) and Poly loss ($\epsilon$=-1). We initialize AvUC with a warmup of 20 epochs and pair Soft-AvUC with Focal loss. Our experiments were conducted on a NVIDIA GeForce RTX 2070 GPU with i7-10700 CPU and NVIDIA RTX A5000 with Intel(R) Xeon(R) Silver 4310 CPU.

\section{Additional Metrics and Results}
\label{extra_metrics}
Apart from the calibration metrics described in the main paper, we further include other calibration metrics and discuss additional results for both synthetic and wild OOD here.
\subsection{Additional Metrics}
\textbf{Maximum Calibration Error (MCE)} \cite{10.5555/2888116.2888120} is similar to ECE, where approximations include binning the predicted probabilities and measuring the maximum absolute difference between the partitioned accuracy and confidence bins. The MCE is well-suited for high-risk applications where the worst case scenarios are to be considered, specifically: MCE $=\max_{\{1 \in ... B\}} |\textrm{acc}(b) - \textrm{conf}(b)|$.

\noindent \textbf{Adaptive ECE (AdaECE)} is proposed to evenly measure samples across bins \cite{nguyen-oconnor-2015-posterior} since the ECE is known to be biased towards bins with higher confidence intervals: AdaECE $= \sum^B_{b=1} \frac{n_b}{N} | \textrm{acc}(b) - \textrm{conf}(b) |$ s.t. $ \forall{b,i} \cdot |B_b| = |B_i| $.

\noindent \textbf{Negative Log-likelihood (NLL)} \cite{hastie01statisticallearning}, also commonly known as CE loss in deep learning. Given a model's probabilities $P_i(y_k|x)$ and the ground truth targets $y_{k}$, the NLL is given as the mean value of the summation across all K classes: NLL $= -\frac{1}{N}\sum^{N}_{i=1} \sum^{K-1}_{k=0} y_k \log{P_i(y_k|x)}$.

\noindent \textbf{Brier Score (BS)} The Brier score \cite{Brier1950VERIFICATIONOF} (BS) is a simple yet important calibration metric which measures the quadratic difference between model calibration and refinement. Specifically, it is defined as the mean of squared differences between the ground truth one-hot vector and the predicted probabilities: BS $= \frac{1}{N}\sum^{N}_{i=1} \sum^{K}_{k=1} |y_k - P_i(y_k|x)|^2$.

In general, classifiers with lower overall calibration scores are considered to be better calibrated. We show the additional results using other calibration metrics for Synthethic OOD in \cref{table:augmented_extra} and Wilds in \cref{table:wilds_extra}. For the NLL and Brier score, we show the errors of misclassified samples as proposed by \cite{mukhoti2020calibrating}. Apart from metrics, calibration behaviour can also be visualized (see \cref{fig:probabilty_densities}) with bin-strength plots and reliability diagrams \cite{NiculescuMizil2005PredictingGP}. For reliability diagrams, perfectly calibrated models will have accuracy bins that equal their confidence.
\subsection{Supplementary OOD results}
In addition to the main calibration metrics (ECE, CECE and KSE) shown in the main paper, we further discuss additional findings using other calibration metrics. We show the results in (\%) scores for MCE and AdaECE computed with 15 bins, along with the misclassified raw error scores for NLL and BS. For both synthetic and Wild OOD, we find that our method generally remains competitive against other loss functions on these additional calibration metrics. \cref{table:augmented_extra}. %This is in contrast to the main paper where the best performance was from the mean constraint form of our method, which had the best results on the main metrics. For additional real-world OOD results, we find that Soft-AvUC performs the best on iWildCam-Wilds, while our method performs best on Camelyon17-Wilds and FmoW-Wilds. This is similar behaviour to synthetic OOD, where our method exhibits competitive performance against the state-of-the-art.

\label{extra_results}
\begin{table*}[!htb]%\setlength\tabcolsep{0.25em}
\centering
\large
\begin{adjustbox}{width=\textwidth}
\begin{tabular}{c|cccc|cccc|cccc}
&\multicolumn{4}{c|}{(a) CIFAR10-C} &\multicolumn{4}{c|}{(b) CIFAR100-C} &\multicolumn{4}{c}{(c) Tiny ImageNet-C}\\
Loss Fn. &MCE &AdaECE &NLL &BS &MCE &AdaECE &NLL &BS &MCE &AdaECE &NLL &BS\\
\hline
CE &34.9$\pm{3.9}$ &11.8$\pm{0.4}$ &3.329$\pm{0.1}$ &0.14$\pm{0.1}$ &24.6$\pm{0.5}$ &9.90$\pm{0.1}$ &3.87$\pm{1.5}$ &0.01$\pm{0.1}$ &40.7$\pm{1.8}$ &15.7$\pm{0.6}$ &0.05$\pm{6.5}$ &0.005$\pm{0.0}$ \\
Focal &22.8$\pm{1.4}$ &7.50$\pm{0.1}$ &2.539$\pm{0.1}$ &0.12$\pm{0.1}$ &27.4$\pm{2.1}$ &13.8$\pm{1.4}$ &3.62$\pm{3.2}$ &0.01$\pm{0.1}$ &37.2$\pm{0.2}$ &13.9$\pm{0.1}$ &0.04$\pm{0.6}$ &0.005$\pm{0.0}$ \\
Inv Focal &32.8$\pm{0.1}$ &12.2$\pm{0.2}$ &3.558$\pm{0.1}$ &0.14$\pm{0.1}$ &23.0$\pm{1.0}$ &9.30$\pm{0.3}$ &3.98$\pm{1.3}$ &0.01$\pm{0.1}$ &43.0$\pm{1.0}$ &16.4$\pm{0.7}$ &0.05$\pm{6.6}$ &0.005$\pm{0.0}$ \\
AvUC &37.4$\pm{0.3}$ &13.2$\pm{0.4}$ &4.465$\pm{0.1}$ &0.15$\pm{0.1}$ &36.2$\pm{0.3}$ &16.1$\pm{0.2}$ &4.82$\pm{1.4}$ &0.01$\pm{0.1}$ &36.8$\pm{2.3}$ &11.3$\pm{0.5}$ &0.04$\pm{3.9}$ &0.005$\pm{0.0}$ \\
Soft AvUC &25.2$\pm{2.8}$ &10.3$\pm{0.2}$ &\textbf{2.425}$\pm{0.1}$ &\textbf{0.11}$\pm{0.1}$ &29.9$\pm{1.1}$ &13.3$\pm{0.1}$ &\textbf{3.33}$\pm{1.6}$ &0.01$\pm{0.1}$ &30.2$\pm{0.1}$ &10.7$\pm{0.1}$ &0.04$\pm{6.0}$ &0.005$\pm{0.0}$ \\
Poly &29.7$\pm{0.1}$ &10.4$\pm{0.4}$ &3.211$\pm{0.1}$ &0.14$\pm{0.1}$ &23.1$\pm{0.3}$ &9.30$\pm{0.3}$ &3.89$\pm{0.4}$ &0.01$\pm{0.1}$ &42.4$\pm{0.3}$ &17.5$\pm{0.2}$ &0.05$\pm{1.2}$ &0.006$\pm{0.0}$ \\
MaxEnt M &24.2$\pm{0.5}$ &6.90$\pm{0.1}$ &2.479$\pm{0.1}$ &0.12$\pm{0.1}$ &20.4$\pm{0.5}$ &11.1$\pm{0.1}$ &3.49$\pm{1.2}$ &0.01$\pm{0.1}$ &29.4$\pm{1.9}$ &10.3$\pm{0.1}$ &0.04$\pm{1.0}$ &0.005$\pm{0.0}$ \\
MaxEnt V &\textbf{19.3}$\pm{0.1}$ &\textbf{6.70}$\pm{0.2}$ &2.531$\pm{0.1}$ &0.12$\pm{0.1}$ &\textbf{17.1}$\pm{0.1}$ &\textbf{7.60}$\pm{0.2}$ &3.57$\pm{4.2}$ &\textbf{0.01}$\pm{0.1}$ &\textbf{26.0}$\pm{0.4}$ &\textbf{9.40}$\pm{0.1}$ &\textbf{0.04}$\pm{1.6}$ &\textbf{0.005}$\pm{0.0}$\\
MaxEnt M+V &32.1$\pm{0.8}$ &9.50$\pm{0.4}$ &2.654$\pm{0.1}$ &0.12$\pm{0.1}$ &23.9$\pm{1.2}$ &12.3$\pm{0.4}$ &3.52$\pm{1.5}$ &0.01$\pm{0.1}$ &29.4$\pm{1.2}$ &10.5$\pm{0.1}$ &0.04$\pm{2.3}$ &0.005$\pm{0.0}$\\ 
\hline
\end{tabular}
\end{adjustbox}
%\vspace{-3mm}
\caption{Additional test scores (\%) for the synthetic OOD datasets computed across different approaches, $\pm$ indicates the standard errors for 3 random seeds, with the best mean scores highlighted in bold.}
\label{table:augmented_extra}
\end{table*}

\begin{table*}[!htb]%\setlength\tabcolsep{0.25em}
\centering
\large
\begin{adjustbox}{width=\textwidth}
\begin{tabular}{c|cccc|cccc|cccc}
&\multicolumn{4}{c|}{(a) Camelyon17-Wilds} &\multicolumn{4}{c|}{(b) iWildCam-Wilds} &\multicolumn{4}{c}{(c) FmoW-Wilds}\\
Loss Fn. &MCE &AdaECE &NLL &BS &MCE &AdaECE &NLL &BS &MCE &AdaECE &NLL &BS\\
\hline
CE &34.2$\pm{0.8}$ &14.4$\pm{1.6}$ &4.10$\pm{0.4}$ &0.80$\pm{0.1}$ &58.7$\pm{1.2}$ &28.6$\pm{0.2}$ &4.50$\pm{0.1}$ &0.01$\pm{0.1}$ &60.4$\pm{0.3}$ &40.3$\pm{0.1}$ &6.10$\pm{0.1}$  &0.01$\pm{0.1}$ \\
Focal &26.6$\pm{2.6}$ &11.8$\pm{1.7}$ &2.60$\pm{0.1}$ &0.70$\pm{0.1}$ &50.9$\pm{2.0}$ &25.4$\pm{1.3}$ &3.80$\pm{0.1}$ &0.01$\pm{0.1}$ &50.7$\pm{0.6}$ &33.9$\pm{0.2}$ &5.10$\pm{0.1}$  &0.01$\pm{0.1}$ \\
Inv Focal &35.5$\pm{0.6}$ &17.0$\pm{2.5}$ &5.70$\pm{0.3}$ &0.80$\pm{0.1}$ &64.2$\pm{1.6}$ &33.2$\pm{2.5}$ &5.50$\pm{0.2}$ &0.01$\pm{0.1}$ &35.4$\pm{0.1}$ &19.1$\pm{0.2}$ &3.90$\pm{0.1}$  &0.01$\pm{0.1}$ \\
AvUC &30.4$\pm{5.9}$ &20.9$\pm{2.4}$ &7.10$\pm{0.6}$ &0.80$\pm{0.1}$ &61.9$\pm{0.1}$ &24.2$\pm{0.8}$ &5.00$\pm{0.1}$  &0.01$\pm{0.1}$ &20.6$\pm{0.3}$ &7.80$\pm{0.1}$ &3.50$\pm{0.1}$  &0.01$\pm{0.1}$ \\
Soft AvUC &27.7$\pm{1.1}$ &18.3$\pm{2.1}$ &1.90$\pm{0.3}$ &0.60$\pm{0.1}$ &55.4$\pm{22.1}$ &\textbf{11.8}$\pm{0.5}$ &\textbf{2.70}$\pm{0.1}$ &\textbf{0.01}$\pm{0.1}$ &14.7$\pm{0.7}$ &9.60$\pm{1.0}$ &3.60$\pm{0.1}$  &0.01$\pm{0.1}$ \\
Poly &33.2$\pm{0.1}$ &12.6$\pm{0.3}$ &4.40$\pm{0.7}$ &0.80$\pm{0.1}$ &60.1$\pm{1.6}$ &28.7$\pm{0.9}$ &4.30$\pm{0.2}$  &0.01$\pm{0.1}$ &30.7$\pm{0.2}$ &16.7$\pm{0.4}$ &3.70$\pm{0.1}$  &0.01$\pm{0.1}$ \\
MaxEnt M &20.6$\pm{9.3}$ &7.00$\pm{3.4}$ &2.00$\pm{0.3}$ &0.60$\pm{0.1}$ &39.4$\pm{4.8}$ &19.9$\pm{3.3}$ &3.90$\pm{0.1}$  &0.01$\pm{0.1}$ &\textbf{10.5}$\pm{0.3}$ &\textbf{5.60}$\pm{0.2}$ &\textbf{3.40}$\pm{0.1}$ &\textbf{0.01}$\pm{0.1}$ \\
MaxEnt V &28.3$\pm{1.3}$ &14.0$\pm{2.2}$ &2.30$\pm{0.1}$ &0.70$\pm{0.1}$ &\textbf{24.8}$\pm{0.6}$ &12.1$\pm{1.0}$ &2.90$\pm{0.1}$  &0.01$\pm{0.1}$ &34.2$\pm{12.8}$ &22.2$\pm{10.0}$ &4.60$\pm{0.7}$ &0.01$\pm{0.1}$ \\
MaxEnt M+V &\textbf{14.8}$\pm{3.1}$ &\textbf{6.00}$\pm{0.8}$ &\textbf{1.40}$\pm{0.1}$ &\textbf{0.50}$\pm{0.1}$ &56.2$\pm{3.1}$ &31.0$\pm{1.2}$ &4.10$\pm{0.1}$  &0.01$\pm{0.1}$ &34.5$\pm{11.8}$ &22.0$\pm{8.4}$ &4.40$\pm{0.6}$ &0.01$\pm{0.1}$ \\
\hline
\end{tabular}
\end{adjustbox}
%\vspace{-3mm}
\caption{Additional test scores (\%) on the real-world OOD Wilds datasets computed across different approaches, $\pm$ indicates the standard errors for 3 random seeds, with the best mean scores highlighted in bold.}
\label{table:wilds_extra}
\end{table*}
\clearpage
\subsection{Effects of Label Smoothing}
Here we discuss the effects of label smoothing on the mean and variance forms of our method. On the left, we show the bin strength densities of our method with constraints placed on the global mean and variance. Then we show the effects of label smoothing which gives better calibrated models with ``flatter'' bin-strength densities. %When including the local constraints, label smoothing only results in a linear shift of the bin densities.

\begin{figure}[!htb]
\small
\includegraphics[width=\columnwidth]{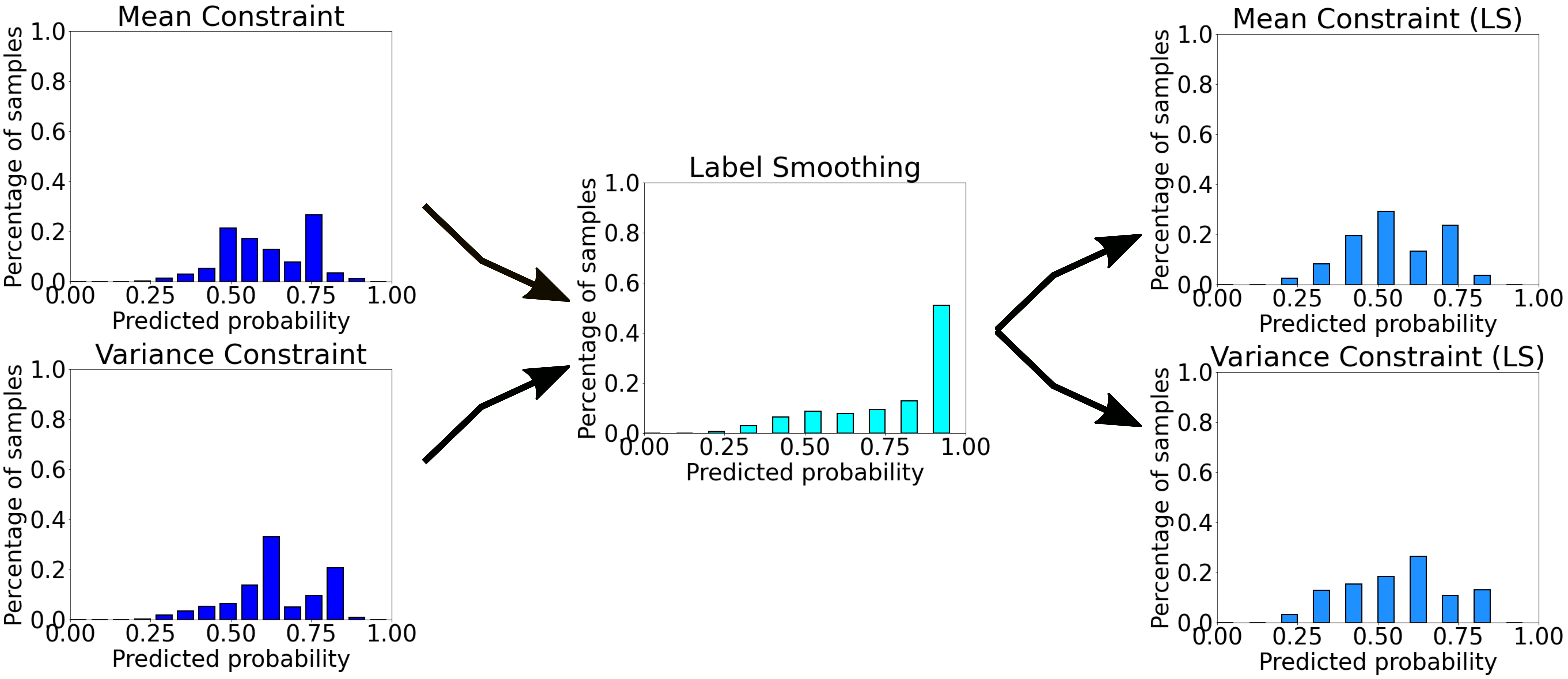}
%\vspace*{-6mm}
\caption{Effects of label smoothing: when combining MaxEnt loss with label smoothing, models are even better calibrated, resulting in``flatter'' bin-strength histograms.}
\label{fig:cifar10_label_smoothing}
\end{figure}

\end{document}